\documentclass{article}

    \PassOptionsToPackage{numbers, compress}{natbib}
    \usepackage[final]{neurips_2025}

\usepackage[utf8]{inputenc} % allow utf-8 input
\usepackage[T1]{fontenc}    % use 8-bit T1 fonts
\usepackage{hyperref}       % hyperlinks
\usepackage{url}            % simple URL typesetting
\usepackage{booktabs}       % professional-quality tables
\usepackage{amsfonts}       % blackboard math symbols
\usepackage{nicefrac}       % compact symbols for 1/2, etc.
\usepackage{microtype}      % microtypography
\usepackage{xcolor}         % colors

\usepackage{amsmath}
\usepackage{adjustbox}
\usepackage{algorithm} 
\usepackage{algpseudocode}
\usepackage{amsthm}
\usepackage{multirow}
\usepackage{colortbl}
\usepackage{subcaption}
\usepackage{graphicx}
\usepackage{wrapfig}
\usepackage{amsthm}
\usepackage{float}

\usepackage{pifont}

\newcommand{\gc}{\cellcolor[rgb]{0.85, 0.92, 0.97}}

\title{PRESTO: Preimage-Informed Instruction Optimization for Prompting Black-Box LLMs}

\author{%
  Jaewon Chu$^1$ \hspace{0.2cm}
  Seunghun Lee$^2$\thanks{Work done while at Korea University.} \hspace{0.2cm}
  Hyunwoo J. Kim$^2$\thanks{Corresponding author}  \\
  $^1$Korea University, $^2$KAIST \\
  \texttt{allonsy07@korea.ac.kr \{llsshh319, hyunwoojkim\}@kaist.ac.kr}\\
}

\begin{document}

\maketitle

\begin{abstract}
Large language models (LLMs) have achieved remarkable success across diverse domains, due to their strong instruction-following capabilities. 
This has led to increasing interest in optimizing instructions for black-box LLMs, whose internal parameters are inaccessible but widely used due to their strong performance.
% Recent approaches leverage white-box LLMs to assist instruction optimization for black-box LLMs by generating instructions from soft prompts.
To optimize instructions for black-box LLMs, recent methods employ white-box LLMs to generate candidate instructions from optimized soft prompts.
However, white-box LLMs often map different soft prompts to the same instruction, leading to redundant queries.
% While previous studies regarded this many-to-one mapping as a redundancy to be avoided, we reinterpret it as useful prior knowledge that can enhance the optimization performance.
While previous studies regarded this many-to-one mapping as a structure that hinders optimization efficiency, we reinterpret it as a useful prior knowledge that can accelerate the optimization.
To this end, we introduce \underline{\textbf{PRE}}image-informed in\underline{\textbf{S}}\underline{\textbf{T}}ruction \underline{\textbf{O}}ptimization (PRESTO), a novel framework that leverages the preimage structure of soft prompts for efficient optimization.
PRESTO consists of three key components: (1) score sharing, which shares the evaluation score with all soft prompts in a preimage; (2) preimage-based initialization, which selects initial data points that maximize search space coverage using preimage information; and (3) score consistency regularization, which enforces prediction consistency within each preimage.
By leveraging preimages, PRESTO achieves the effect of effectively obtaining 14 times more scored data under the same query budget, resulting in more efficient optimization.
Experimental results on 33 instruction optimization tasks demonstrate the superior performance of PRESTO.
Code is available at \href{https://github.com/mlvlab/PRESTO}{https://github.com/mlvlab/PRESTO}.
\end{abstract}

\section{Introduction}
\label{sec:intro}
Large language models (LLMs) have demonstrated strong performance across a wide range of domains~\cite{grattafiori2024llama, achiam2023gpt, guo2025deepseek, thirunavukarasu2023large, naveed2023comprehensive}.
This success is largely attributed to their impressive instruction-following capabilities, which have led to growing interest in discovering effective instructions to enhance their performance~\cite{ouyang2022training, liu2023pre}.
In particular, LLMs provided through APIs (\textit{i.e.,} black-box LLMs), such as GPT-4~\cite{achiam2023gpt}, are widely used and show exceptionally strong performance.
However, optimizing instructions for the black-box LLMs is a challenging problem, since their internal parameters are inaccessible.
To tackle this challenge, recent studies have explored various strategies for optimizing instructions for black-box LLMs, without access to internal model parameters~\cite{zhou2022large, yang2023large, guo2024connecting, fernando2024promptbreeder, pryzant2023automatic, shi2024efficient}.

\begin{figure}[t]
    \centering
    \begin{subfigure}[h]{0.38\linewidth}
        % \vspace{0.4em}
        \includegraphics[width=0.97\textwidth]{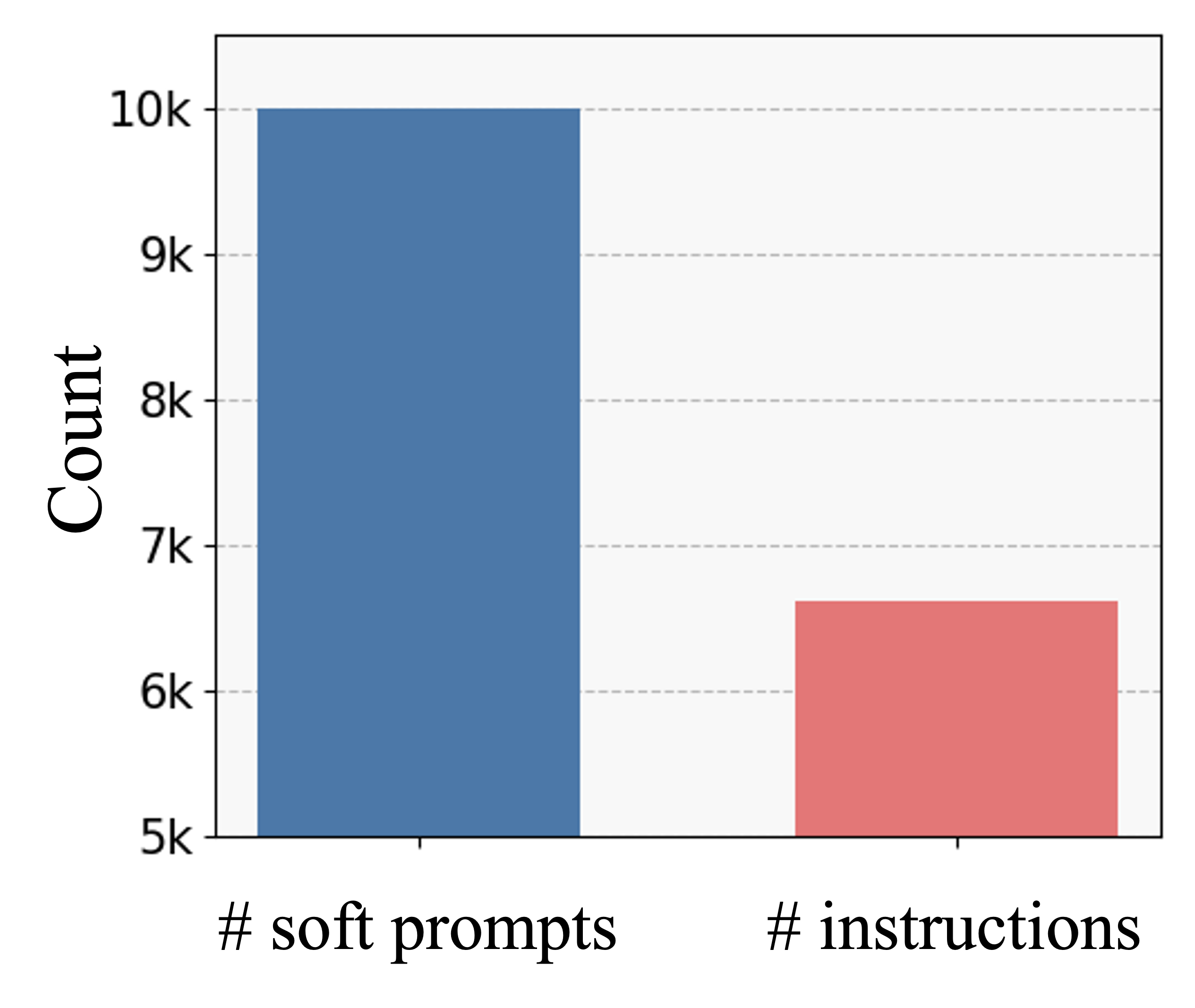}
        \caption{} 
        \label{subfig:unique_instruction}
    \end{subfigure}
    \begin{subfigure}[h]{0.60\linewidth}
        % \vspace{-0.5em}
        \includegraphics[width=1.0\textwidth]{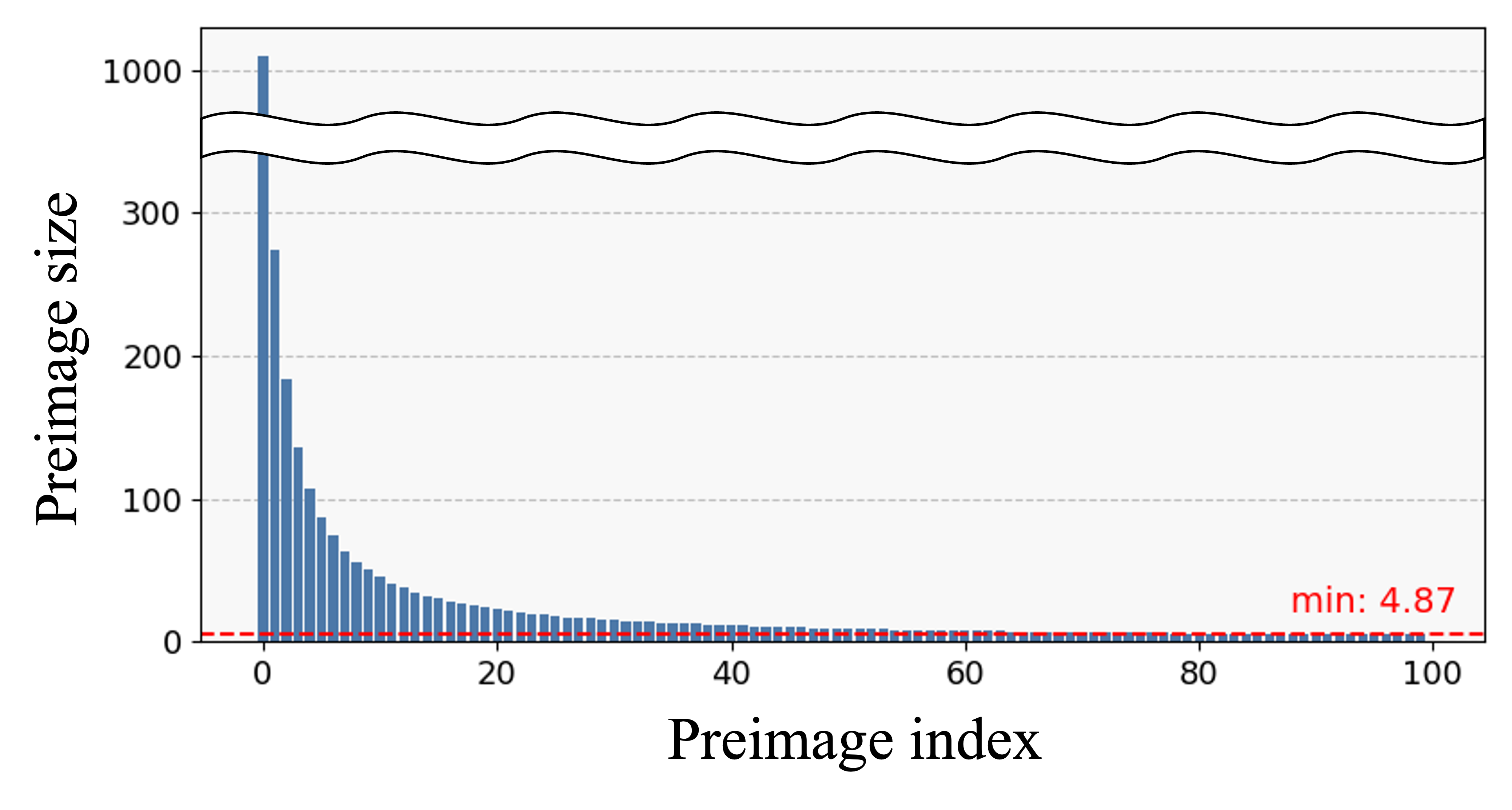}
        \vspace{-1.5em}
        \caption{} 
        \label{subfig:group_size}
    \end{subfigure}
    \caption{
    Motivating observations illustrating the many-to-one mapping from soft prompts to instructions in a white-box LLM (LLaMA3.1-8B-Instruct~\cite{grattafiori2024llama}). 
    Figure~\ref{subfig:unique_instruction} shows that the white-box LLM produces approximately 6,500 unique instructions from 10,000 distinct soft prompts.
    Figure~\ref{subfig:group_size} presents the distribution of preimage sizes, displaying the top 100 largest preimages. 
    The largest preimage contains more than 1,000 soft prompts, while the 100th largest has around 5.
    Both figures report the average experimental results over the instruction induction tasks used in Table~\ref{tab:main results}.}
    % \vspace{-50pt}
    \label{fig:motivation_observation}
\end{figure}

Recently, some studies~\cite{chen2024instructzero, hu2024localized, lin2024use} have leveraged open-source LLMs (\textit{i.e.,} white-box LLMs)~\cite{grattafiori2024llama, jiang2024mixtral, yang2024qwen2} to assist instruction optimization for black-box LLMs, demonstrating promising results and attracting growing interest.
Specifically, these methods optimize a soft prompt, which is taken as input to the white-box LLM.
The optimization is performed using black-box optimization algorithms such as Bayesian Optimization~\cite{frazier2018tutorial, shahriari2015taking} or Neural Bandits~\cite{zhou2020neural, zhang2020neural}, guided by a score predictor regression model, allowing the white-box LLM to generate effective instructions for black-box LLMs.
However, as shown in Figure~\ref{subfig:unique_instruction}, white-box LLMs often generate identical instructions from distinct soft prompts.
% It leads to repeatedly querying soft prompts that yield the same outputs during the optimization process, which ultimately reduces query efficiency.
It leads to repeatedly querying soft prompts that yield the same outputs during the optimization process, which ultimately hinders the optimization process by reducing query efficiency.
To avoid redundant queries, previous studies either sample soft prompts that are well-separated in the soft prompt space~\cite{lin2024use} or filter soft prompts that generate distinct instructions~\cite{hu2024localized}.

% \begin{figure}[t]
%     \centering
%     \includegraphics[width=0.9\textwidth]{Figures/motivation.png}
%     \caption{} 
%     \label{fig:motivation_observation}
% \end{figure}Many-to-one mapping from soft prompts to instructions

% While previous studies have treated the generation of identical instructions from different soft prompts (\textit{i.e.,} many-to-one structure) as a redundancy to be mitigated, we reinterpret this as a valuable structure that can facilitate the optimization process.
While previous studies have treated the generation of identical instructions from different soft prompts (\textit{i.e.,} many-to-one structure) as a redundancy that hinders optimization, we reinterpret this as a valuable structure that can facilitate the optimization process.
Specifically, the set of soft prompts that generate the same instruction forms the preimage of that instruction under the white-box LLM.
This preimage imposes a strong inductive bias over the search space: all soft prompts within a preimage share the same objective function value.
Since we follow previous settings~\cite{lin2024use, hu2024localized} that sample a sufficiently large set of $N$ soft prompts and search for the optimal solution within them, we do not observe the full preimage, but only a subset of it.
We refer to such subsets as preimages throughout the paper, and provide the size distribution of these preimages in Figure~\ref {subfig:group_size}.

Building on this insight, we propose PRESTO, a novel instruction optimization framework that explicitly leverages the many-to-one structure to facilitate instruction optimization for black-box LLMs.
PRESTO consists of three components.
First, we present the score-sharing method, where once the score is evaluated through the black-box LLM, it is shared with all soft prompts within a preimage.
This effectively enlarges the amount of scored data without additional calls to the black-box LLM.
Second, we introduce preimage-based initialization, where we select the initial soft prompts regarding the preimage information so that they cover the search space maximally.
Finally, we propose score consistency regularization, which adds a regularization term to encourage the score predictor to predict identical scores for soft prompts within the same preimage. % 여기서 갑자기 score predictor이야기가 나와도 되나?
We evaluate the instruction optimization performance of PRESTO on 30 instruction induction tasks and three arithmetic reasoning tasks, and achieve state-of-the-art performance compared to existing baselines.

The main contributions of our work are:
\begin{itemize}
    \item We reinterpret the many-to-one structure between the soft prompts and instruction, previously viewed as a challenge, as a rich informative structure that facilitates instruction optimization for black-box LLMs.
    \item Leveraging this insight, we introduce PRESTO, a novel framework that consists of score sharing, preimage-based initialization, and score consistency regularization.
    \item PRESTO achieves state-of-the-art performance across 30 instruction induction and 3 arithmetic reasoning tasks.
\end{itemize}
\section{Related Works}
\noindent\textbf{Instruction Optimization for Black-box LLMs}
Instruction optimization has been widely explored as a way to improve the performance of large language models (LLMs) on downstream tasks~\cite{opsahl2024optimizing, lin2024prompt}.
In particular, when using black-box LLMs such as GPT-4~\cite{achiam2023gpt}, where access to model parameters is restricted, optimization methods rely on model outputs to guide the search for better instructions. 
Under this setting, various approaches have been proposed, including evolutionary algorithms~\cite{guo2024connecting, fernando2024promptbreeder}, LLM-driven meta-optimization~\cite{zhou2022large, yang2023large}, and bandit-style or heuristic search methods~\cite{shi2024efficient, pryzant2023automatic}. 
These works demonstrate that instruction quality can be improved even without access to gradients or internal representations by querying the black-box model efficiently.

More recently, some methods~\cite{chen2024instructzero, hu2024localized, lin2024use}  incorporate open-source white-box LLMs~\cite{grattafiori2024llama, yang2024qwen2, jiang2024mixtral, jiang2023mistral7b} to assist the optimization process. 
Rather than optimizing instruction texts directly, they optimize soft prompts, which are continuous embeddings that the white-box model maps into instructions.  
InstructZero~\cite{chen2024instructzero} leveraged Bayesian Optimization~\cite{lee2023advancing, chuinversion, leelatent} to search for the optimal soft prompts for black-box LLM.
INSTINCT~\cite{lin2024use} leveraged NeuralUCB~\cite{zhou2020neural} with an LLM-based score predictor, which was the first to point out the many-to-one schema and approached it indirectly by sampling soft prompts to be well-separated.
And ZOPO~\cite{hu2024localized} proposed a zeroth-order optimization algorithm~\cite{chen2017zoo} for local search, which addresses this redundancy by simply discarding all but one soft prompt that produces the same instruction.
In contrast, we retain all soft prompts by introducing preimages and facilitate the optimization.
\section{Preliminaries}
\label{sec:prelim}
\noindent\textbf{Problem Formulation}
\label{subsec:problem_formulation}
% Instruction optimization aims to find a human-readable instruction $v$ that maximizes the expected performance of a black-box LLM $f$ (e.g., ChatGPT) on a validation set $\mathcal{D}_{\text{val}} = \{(x_i, y_i)\}_{i=1}^n$, where $v$ is provided with each input $x_i$.
Instruction optimization aims to find an instruction $v$ that guides a language model to perform a given task effectively. 
To be specific, the goal is to find the instruction $v$ that maximizes the task-specific score function $h$ by guiding a black-box LLM $f_\text{b}$ to generate the correct answer $y$, which is formally given as:
\begin{equation}
\label{eq:IO}
    v^* = \underset{v \in \Omega}{\arg\max}\ \mathbb E_{(x,y) \in {D_{\text{val}}}}\big[h(f_{\text{b}}(v, x), y)\big],
\end{equation}
% where $\Omega$ is the discrete sequence domain and $\mathcal{D}_{\text{val}} = \{(x_i,y_i)\}_{i=1}^M$ is a validation set.
where $\mathcal{D}_{\text{val}} = \{(x_i,y_i)\}_{i=1}^M$ is a validation set, and $\Omega$ denotes the search space of instructions, typically a discrete sequence domain (\textit{e.g.,} natural language prompts or token sequences).
However, directly searching over discrete instruction sequences is challenging, as it constitutes a combinatorial optimization problem over the space of token configurations.
To address this, InstructZero~\cite{chen2024instructzero} reformulates the discrete instruction search as a continuous optimization problem by leveraging a white-box LLM $f_\text{w}$.
Specifically, it optimizes a soft prompt $z \in \mathbb{R}^{N_z \times d}$, where $N_{z}$ is the number of tokens and $d$ is the embedding dimension, to generate the optimal instruction $v^\ast$.
The soft prompt is concatenated with the token embeddings of input-output exemplars $E=\{(x_i,y_i)\}_{i=1}^\kappa$ and fed into the white-box LLM $f_\text{w}$, which then generates an instruction $v = f_\text{w}(z, E)$.
Formally, the instruction optimization problem is defined as:
\begin{equation}
\label{eq:discreteIO}
    z^* = \underset{z \in \mathcal{Z}}{\arg\max}\ \mathbb E_{(x,y) \in {D}_{\text{val}}}\big[h(f_{\text{b}}(f_\text{w}(z, E), x), y)\big],
\end{equation}
where $\mathcal{Z}$ is the soft prompt space.
% The instruction generated from the optimized soft prompt $z^*$ is subsequently evaluated on a held-out test set ${D}_{\text{test}}$.
In this formulation, we optimize $z$ to find the optimal instruction $v^*$ that maximizes the expected value of the score function $h$.
Once the optimal soft prompt $z^*$ is obtained, the corresponding instruction $v^*$ is generated by the white-box LLM $f_\text{w}$, \textit{i.e.,} $v^*=f_w(z^\ast,E)$ and subsequently evaluated on a held-out test set ${D}_{\text{test}}$.
Since the exemplars $E$ are fixed for each task, we omit them from the notation in the rest of our paper.
Following previous works~\cite{chen2024instructzero,hu2024localized, lin2024use}, we assume that both the white-box LLM $f_\text{w}$ and the black-box LLM $f_\text{b}$ are deterministic.
%($\textit{i.e.,}$ decoding temperature is set to 0)

\noindent\textbf{LLM-based Score Predictor for Instruction Optimization.}
\label{subsec:LLM-based score predictor}
Our method builds upon INSTINCT~\cite{lin2024use}, which employs a frozen white-box LLM as a feature extractor to predict the score of soft prompt, and uses a NeuralUCB~\cite{zhou2020neural} for instruction optimization.
Given a soft prompt $z$, the white-box LLM produces an embedding $g(z)$, the last token representation of the final transformer layer.
This embedding is then passed to a score predictor $m(g(z); \theta)$ (\textit{e.g.,} an MLP), which predicts the performance of the instruction generated from $z$, \textit{i.e.},  $m(g(z); \theta) \approx \mathbb E_{(x,y) \in {D}} [h(f_\text{b}(f_\text{w}(z), x), y) ]$.
At each optimization step, the score predictor $m(\cdot; \theta)$ is trained on previously evaluated soft prompts and their corresponding scores, and selects the next query that maximizes the upper confidence bound.
We provide further details of NeuralUCB in the supplement.
Since computing $g(z)$ requires a full forward pass through the LLM, INSTINCT mitigates this cost by precomputing the embeddings of a candidate soft prompt set $Z = \{z_i\}_{i=1}^N$ at the beginning of the optimization, which is sampled using a quasi-random method.
To this end, the instruction optimization task is reduced to searching for the best solution within the precomputed embedding set, as the white-box LLM is frozen during the optimization process.
% This candidate set $Z$ is constructed by sampling low-dimensional vectors using a quasi-random method that evenly covers the space with randomness.

% \noindent\textbf{NeuralUCB-based Soft Prompt Selection.} 
% At each optimization step, the score predictor $m(g(z, E), \theta)$ is trained on previously evaluated soft prompts and their corresponding scores.
% The model's predicted score $\mu(z)$ and its associated uncertainty $\sigma(z)$ are computed as:
% \begin{align}
%     \mu(z) &= m(g(z,E), \theta), \\
%     \sigma(z) &= \sqrt{
%         \nabla_\theta m(g(z,E), \theta)^\top V^{-1} \nabla_\theta m(g(z,E), \theta)
%     }, \\
%     \text{where} \quad V &= \sum_{\tau=1}^{t} \nabla_\theta m(g(z_\tau, E), \theta) \nabla_\theta m(g(z_\tau,E), \theta)^\top + \lambda I,
% \end{align}
% where $\lambda$ is a regularization coefficient and $t$ is the number of observed data.
% The next prompt to evaluate is selected by maximizing an Upper Confidence Bound (UCB):
% \begin{equation}
%     z_{\text{next}} = \underset{z \in Z}{\arg\max}\  \mu(z) + \beta^{1/2} \sigma(z),
% \end{equation}
% where $\beta$ is a weighting parameter that balances exploration and exploitation.
\section{Method}
\label{sec:method}
\begin{figure}[t]
    \centering
    \includegraphics[width=\textwidth]{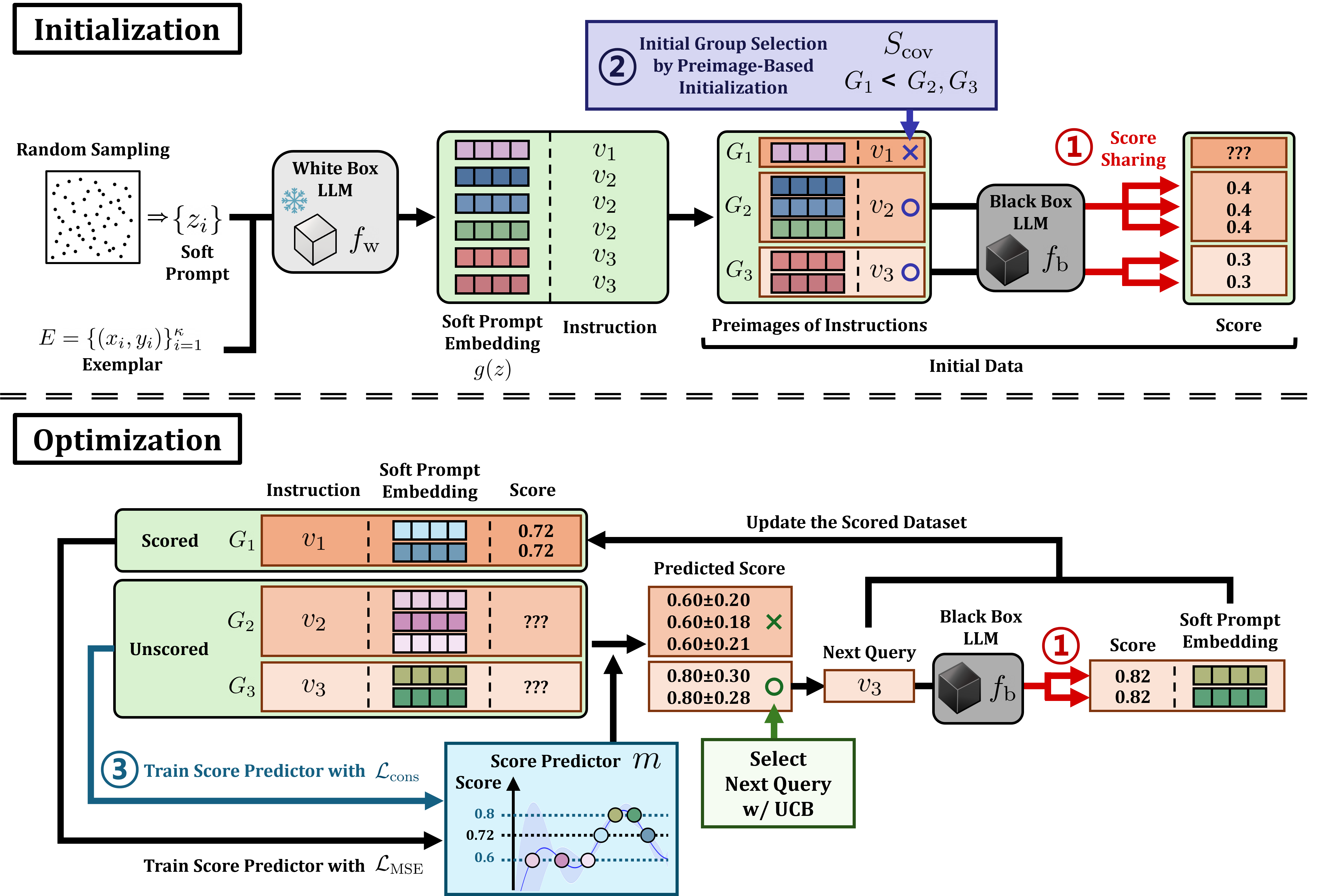}
    \caption{\textbf{The overall process of our proposed PRESTO framework}. It consists of two main stages: initialization and optimization. 
    In the initialization stage, our method performs \ding{172} \emph{preimage-based score sharing} (Section~\ref{main_sec:ScoreAssignment}) and \ding{173} \emph{preimage-based initialization} to improve search space coverage (Section~\ref{main_sec:GroupInitialization}). 
    For the optimization stage, we train the score predictor with \ding{174} \emph{score consistency regularization} (Section~\ref{main_sec:NeuralUCB}) and we apply \ding{172} \emph{preimage-based score sharing} to share scores of newly observed data within the same preimage.}
    \label{fig:main_fig}
\end{figure}

In this section, we propose \underline{\textbf{PRE}}image-informed in\underline{\textbf{S}}\underline{\textbf{T}}ruction \underline{\textbf{O}}ptimization (PRESTO) which is a novel instruction optimization framework that leverages the many-to-one mapping between soft prompts $z \in Z\subset \mathcal{Z}$ and instructions $v \in \Omega$ (or the preimages of instructions, which is defined in Section~\ref{main_sec:ScoreAssignment}) as prior knowledge to facilitate more efficient optimization.
We first introduce a score sharing method that shares the score value of one soft prompt with all other soft prompts in the same preimage, effectively enlarging the scored data without additional evaluations of black-box LLM $f_\text{b}$.
Next, we present a preimage-based initialization method designed to maximize coverage of the search space under score sharing.
Finally, we propose a score consistency regularization that leverages preimage information as prior knowledge to encourage the score predictor to predict identical scores for soft prompts belonging to the same preimage.
We provide the overall framework of our PRESTO in Figure~\ref{fig:main_fig}.

\subsection{Preimage-Based Score Sharing}
\label{main_sec:ScoreAssignment}
During the instruction optimization, we observe that the white-box LLM $f_\text{w}$ often generates identical instructions from distinct soft prompts, \textit{i.e.}, $f_\text{w}(z) = f_\text{w}(z')$, leading to the same score value.
This redundancy leads to unnecessary queries during optimization, hindering the efficiency of instruction optimization.
While previous works treated this redundancy as an obstacle to efficient optimization, we instead leverage this information as prior knowledge about the objective function to facilitate optimization.
To this end, we propose a simple score sharing scheme that associates a large number of soft prompts with a score value without the additional evaluations of a black-box LLM $f_\text{b}$.

Our goal is to share the score of an evaluated soft prompt $z$ with other soft prompts that generate the same instruction.
To enable this score sharing, we first define the \textit{preimage} of each instruction which consists of all soft prompts that map to the same instruction under the white-box model $f_\text{w}$.
Establishing this preimage structure requires two steps.
First, we sample a soft prompt set $Z = \{z_i\}_{i=1}^N$ using a quasi-random method~\cite{morokoff1994quasi, renardy2021sobol}, which is a widely adopted method to sample the data points that evenly cover the soft prompt space~\cite{chen2024instructzero, lin2024use, hu2024localized}.
Assuming that the soft prompt set size $N$ is large enough to represent the soft prompt space $\mathcal{Z}$, the original optimization problem defined in Eq.~\eqref{eq:discreteIO} reduces to searching for the best solution among the set of $N$ data points, denoted by $Z \subset \mathcal{Z}$.

Next, for each soft prompts $z_j \in Z$, we generate the set of instructions $V = \{v_i\}_{i=1}^M$, using the white-box LLM $f_\text{w}$:
\begin{align}
    {V} = \{v_i\}_{i=1}^{M} = \{f_\text{w}(z_j)\ |\ j=1,\dots,N\}.
\end{align}
Since the different soft prompts often generate the identical instruction (\textit{i.e.,} many-to-one mapping), the number of instructions $M=|V|$ is smaller than or equal to $N$.
The construction of $Z$ and $V$ is performed only once before the optimization process begins.

With the soft prompt set $Z$ and the corresponding instruction set $V$, we now define the preimage of each instruction.
The preimage of an instruction $v$ is the set of soft prompts in $Z$ that generate $v$ under the white-box model $f_\text{w}$:
\begin{align}
    f^{-1}_w(v) = \{z\in Z \ |\ f_\text{w}(z) = v \}.
\end{align}
This preimage contains all soft prompts in $Z$ that generate $v$, and will serve as the basis for score sharing.
Once the preimages $f_\text{w}^{-1}(v)$ for all $v \in V$ are established, we apply score sharing across soft prompts that belong to the same preimage during the optimization.
Specifically, after querying the black-box model $f_\text{b}$ with an instruction $v \in V$, we obtain a score of the instruction.
This score is then shared to all soft prompts in the preimage $f_\text{w}^{-1}(v)$.
By sharing scores in this manner, we effectively enlarge the training data for the score predictor $m(g(z); \theta)$ without additional calls to the black-box LLMs.
Moreover, score sharing avoids redundant evaluations of soft prompts that lead to the same instruction and improves optimization efficiency.
% Score sharing method not only improves the robustness of the score prediction, but also enhances optimization efficiency by avoiding redundant queries during the optimization.
% Once the instruction $v_i$ is evaluated via the black-box LLM $f_{\text{b}}$, all embeddings in the group $G_i$ share the same score.
% This effectively enlarges the training set for the score predictor $m(g(z), \theta)$, improving its ability to estimate the score.

\subsection{Preimage-Based Initialization for Maximizing Search Space Coverage}
\label{main_sec:GroupInitialization}
Here, we introduce a preimage-based initialization method that selects initial data points based on the preimage information defined in Section~\ref{main_sec:ScoreAssignment}.
At the beginning of the optimization, the score predictor $m(g(z); \theta)$ (Section~\ref{sec:prelim}) is trained on the initial dataset, and its predictions are used to select the next data points to query the black-box LLM $f_\text{b}$.
In black-box optimization, it is well known that broadly covering the search space at initialization is crucial for effective optimization~\cite{jones1998efficient, eriksson2019scalable,daulton2020differentiable,daulton2021parallel}.
Our score sharing method introduced in Section~\ref{main_sec:ScoreAssignment} expands the initial dataset without additional queries to the black-box LLM $f_\text{b}$, enabling a more sample-efficient initialization.
To further enhance the search space coverage, we propose a preimage-based initialization method that complements score sharing by promoting a broader initial data distribution.
% While the score-sharing method from the previous section expands the initial dataset by sharing a score with all soft prompts in preimage without additional queries to the black-box LLM $f_\text{b}$, the expanded dataset (\textit{i.e.,} all score-shared soft prompts in selected preimage) often biases toward specific regions of the search space.
% This bias hinders the initial data from representing the search space.

To this end, we design a coverage score $S_{\text{cov}}$ to guide the selection of an initial preimage set $G^{\text{init}}$ that maximally covers the entire set of soft prompt embeddings $G^{\text{total}} = \{g(z) \mid z \in Z\}$.
We conduct initialization in the embedding space rather than the raw soft prompt space, since the optimization operates over the soft prompt embeddings.
These embeddings are precomputed and remain fixed throughout the optimization, as described in Section~\ref{sec:prelim}.
For each instruction $v_i$, we define its corresponding preimage group in the embedding space as $G_i = \{g(z) \mid z \in f^{-1}_\text{w}(v_i)\}$.

Since finding the optimal combination of $N_{\text{init}}$ preimages that maximizes the coverage score $S_{\text{cov}}$ is a computationally intractable combinatorial optimization problem, we adopt a greedy algorithm to iteratively select one preimage at a time.
Specifically, the coverage score $S_{\text{cov}}$ consists of two components: the representativeness score $S_{\text{rep}}$ and the size score $S_{\text{size}}$. 
The representativeness score $S_{\text{rep}}$ encourages the selection of a preimage group $G_i$ that, when combined with already selected preimage groups $G^{\text{init}}$, most closely matches the distribution of the candidate set $G^{\text{total}}$, defined as:
\begin{align}
    S_{\text{rep}}(G_i;G^{\text{init}}, G^{\text{total}}) = 1 -
    \frac{\text{MMD}^2(G_i \cup G^{\text{init}}, G^{\text{total}})}{\max_j \text{MMD}^2(G_j \cup G^{\text{init}}, G^{\text{total}})},
\end{align}
where the MMD$^2$ is the squared Maximum Mean Discrepancy. 
MMD$^2$ is a widely used metric to estimate the similarity between two sets, which is defined as:
\begin{align}
\text{MMD}^2(X, Y) = \mathbb{E}_{x, x' \sim X} [k(x, x')] + \mathbb{E}_{y, y' \sim Y} [k(y, y')] 
 - 2\mathbb{E}_{x \sim X, y \sim Y} [k(x, y)]
\end{align}
where $k(\cdot, \cdot)$ is a positive definite kernel.
To densely cover the search space, we propose the size score $S_{\text{size}}$, which is defined as relative preimage size: $S_{\text{size}}(G_i) = |G_i|\big/{\max_j |G_j|}$.
% It rewards the larger preimage, since larger preimages contribute more data and thus provide richer information for the objective function.
% To densely cover the search space, we propose the size score $S_{\text{size}}$. 
% It rewards the larger preimage, since larger preimages contribute more data and thus provide richer information for the objective function.
% It is defined as follows:
% \begin{align}
%     S_{\text{size}}(G_i) = |G_i|\big/{\max_j |G_j|}.
% \end{align}
Combining the two scores, we define the coverage score for the $G_i$:
\begin{align}
    S_{\text{cov}}(G_i;G^{\text{init}}, G^{\text{total}}) 
    = 
    S_{\text{size}}(G_i) + S_{\text{rep}}(G_i; G^{\text{init}}, G^{\text{total}}).
\end{align}
Starting from an empty set $G^{\text{init}}$, we iteratively select the preimage with the highest coverage score $S_{\text{cov}}$ and add it to $G^{\text{init}}$ until the number of initial preimages reaches $N_{\text{init}}$. This initialization maximizes the coverage of the candidate set $G^{\text{total}}$.
We provide the visualization to demonstrate the effectiveness of our initialization method in Section~\ref{analysis:Preimage_initialization}.

\subsection{Score consistency regularization for score predictor}
\label{main_sec:NeuralUCB}
\begin{figure}[t]
    \centering
    \begin{subfigure}[h]{0.27\linewidth}
        \includegraphics[width=\textwidth]{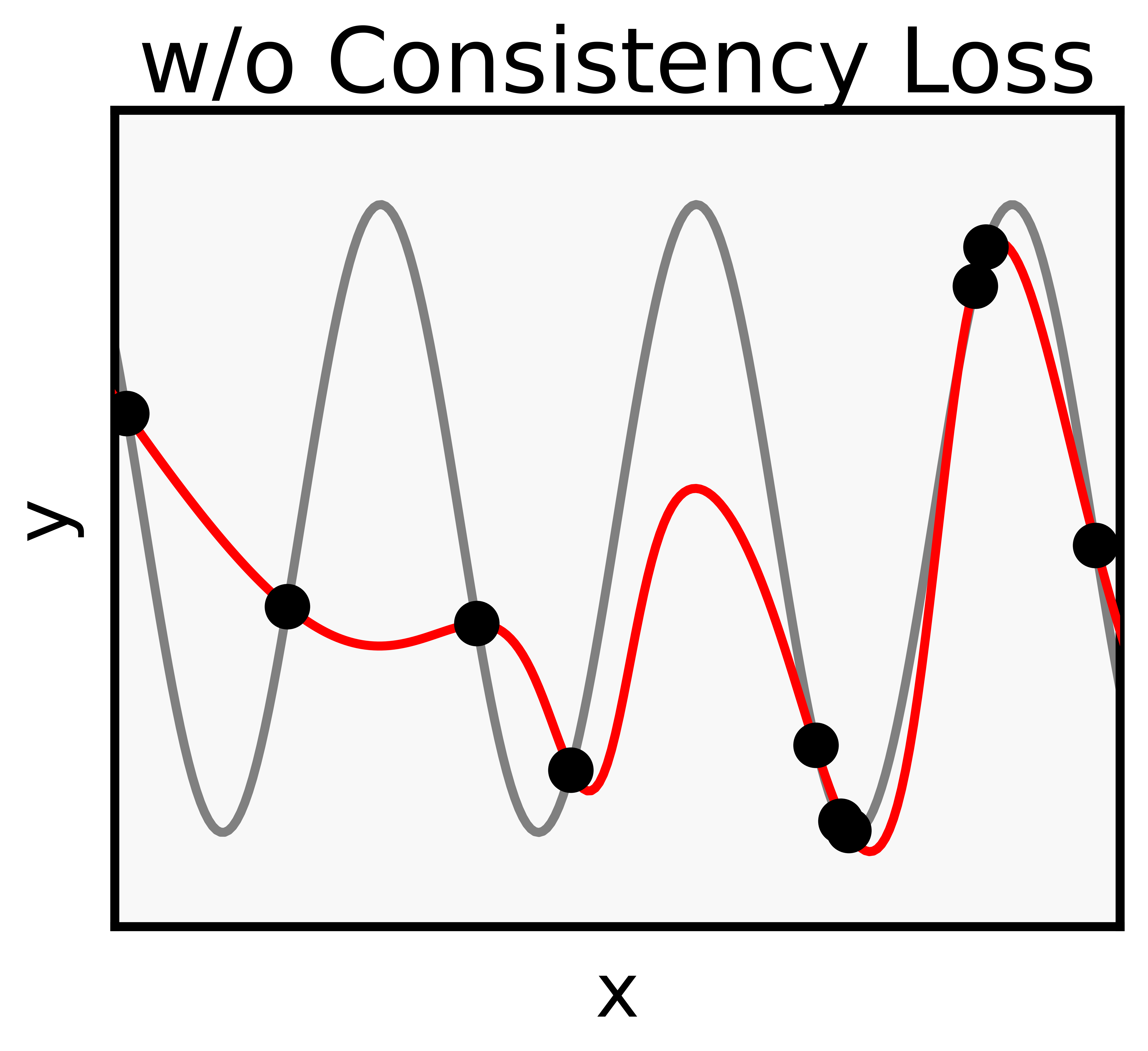}
        \caption{Train with $\mathcal{L}_{\text{MSE}}$ only.} 
        \label{subfig:toy_wo_consistency}
    \end{subfigure}
    \hspace{10pt}
    \begin{subfigure}[h]{0.27\linewidth}
        \includegraphics[width=\textwidth]{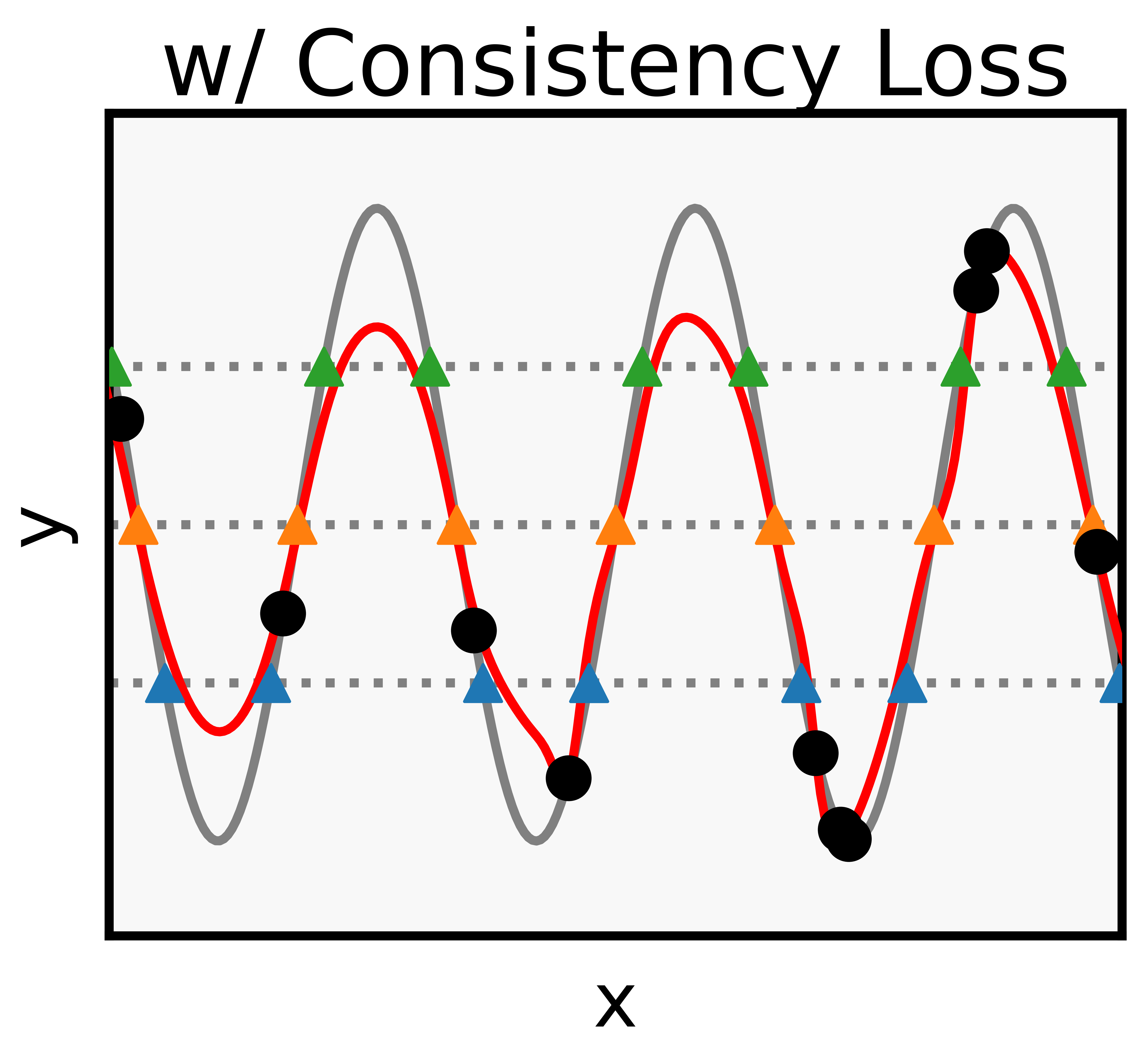}
        \caption{Train with $\mathcal{L}_{\text{MSE}}+\gamma\mathcal{L}_{\text{cons}}$.} 
        \label{subfig:toy_w_consistency}
    \end{subfigure}
    \hspace{5pt}
    \begin{subfigure}[h]{0.25\linewidth}
        \includegraphics[width=\textwidth]{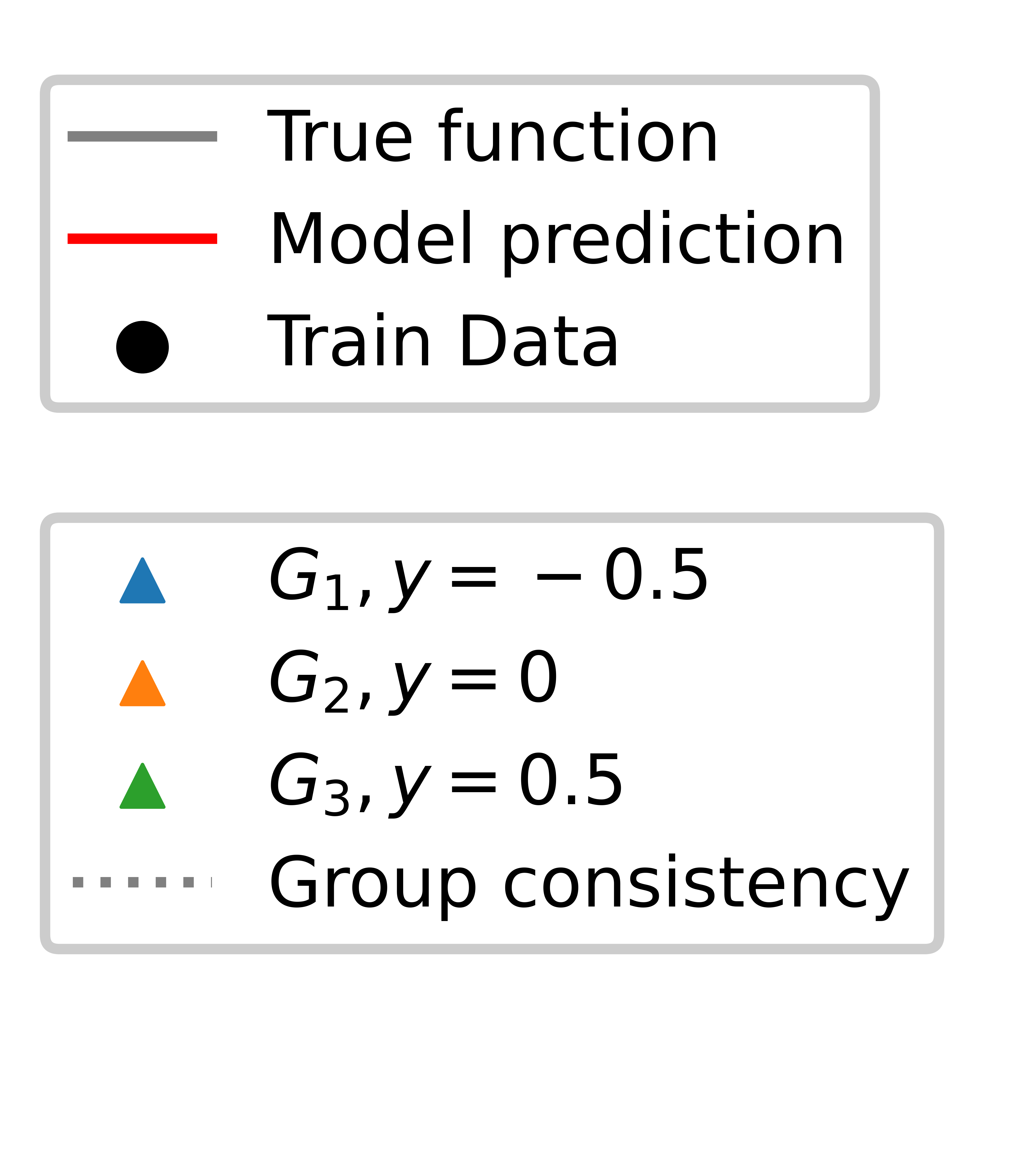}
    \end{subfigure}
    \caption{\textbf{Toy example} comparing models trained w/o and w/ our consistency loss $\mathcal{L}_{\text{cons}}$ in Eq.~\eqref{eq:consistency_loss}.} 
    \label{fig:toy_consistency}
    % \vspace{-5pt}
\end{figure}

Here, we propose a score consistency regularization that encourages the score predictor $m(g(z); \theta)$ to produce the same prediction for all soft prompts in preimages that have not been evaluated by the black-box function.
During the optimization, the score predictor is trained with the scored data to predict the score of each soft prompt in the candidate set $Z$ and estimate its uncertainty for selecting the next query to evaluate.
Leveraging the score sharing method defined in Section~\ref{main_sec:ScoreAssignment} informs the score predictor that data points within the same preimage share identical scores in a supervised manner.
However, since the score predictor lacks information about score consistency within unscored preimages, it is unable to make consistent predictions for data points in these unscored preimages.
It often hinders the score predictor from predicting the ground truth score and selecting high-scored data. 

To ensure consistent predictions within each unscored preimage, we propose a score consistency regularization term $\mathcal{L}_{\text{cons}}$, which is defined as:
% \begin{align}
%     \mathcal{L}_{\text{cons}} = 
%     \frac{1}{K}\sum_{k=1}^K
%     \frac{1}{\binom{|f^{-1}_\text{w}(v_k)|}{2}}
%     \sum_{\substack{z_i, z_j \in f^{-1}_\text{w}(v_k)\\ i<j }}
%     \left(m(g(z_i);\theta) - m(g(z_j);\theta)\right)^2,
%     \label{eq:consistency_loss}
% \end{align}
% where $K$ denotes the number of unscored preimages.
\begin{align}
    \mathcal{L}_{\text{cons}} = 
    \mathbb E_{v\in V_{\text{unseen}}}
    \mathbb E_{z, z' \in f^{-1}_\text{w}(v)}
    \left|m(g(z);\theta) - m(g(z');\theta)\right|^2,
    \label{eq:consistency_loss}
\end{align}
where $V_\text{unseen} \subset V$ denotes the set of instructions that has not been evaluated by the black-box LLM $f_b$.
We note that $\mathcal{L}_{\text{cons}}$ is an unsupervised loss.
While the consistency regularization includes pairwise terms per preimage group, each unscored preimage size is not excessively large in practice, so the computation remains tractable. 
The final loss for training the score predictor model is given by:
\begin{align}
    \mathcal{L} = \mathcal{L}_{\text{MSE}} + \gamma \mathcal{L}_{\text{cons}},
\end{align}
where $\mathcal{L}_{\text{MSE}}$ is the mean squared error loss computed over the scored preimages, and $\gamma$ is a hyperparameter controlling the strength of the regularization.
To avoid premature convergence to incorrect predictions, we employ a simple linear scheduling strategy as $\gamma(t) = \gamma_{\text{max}}\cdot \min\big(1, {t}/{T} \big)$, where $t$ represents the current epoch and $T$ is a warm-up duration.
This schedule allows the score predictor $m(g(z);\theta)$ to learn accurate patterns from the scored data and gradually incorporate the score equality constraint of unscored data.

Figure~\ref{fig:toy_consistency} shows a toy example illustrating the effect of the proposed consistency loss.
We use a simple model with two linear layers. 
In Figure~\ref{subfig:toy_wo_consistency}, the model is trained only with the $\mathcal{L}_{\text{MSE}}$ on the scored data, while in Figure~\ref{subfig:toy_w_consistency}, $\mathcal{L}_{\text{cons}}$ is additionally applied to unscored data.
We assume there are three unscored preimages, each represented by a different marker shape.
Although the model is only given the information that data points within each preimage share the same score, the $\mathcal{L}_{\text{cons}}$ allows it to make more accurate predictions on the unscored data.

\section{Experiments}
\subsection{Experimental settings}
\label{subsec:experimental_setting}
We evaluate our proposed method, PRESTO, on 30 instruction induction tasks~\cite{honovich2023instruction}, a benchmark widely used to assess instruction optimization performance, and 3 arithmetic reasoning tasks~\cite{cobbe2021training, ling2017program, patel2021nlp}.
We compare PRESTO with six competitive instruction optimization baselines: APE~\cite{zhou2022large}, InstructZero~\cite{chen2024instructzero}, INSTINCT~\cite{lin2024use}, EvoPrompt~\cite{guo2024connecting}, ZOPO~\cite{hu2024localized}, and OPRO~\cite{yang2023large}.
We use LLaMA3.1-8B-Instruct~\cite{grattafiori2024llama} as the white-box LLM $f_\text{w}$ to generate candidate instructions, and GPT-4.1 as the black-box model $f_\text{b}$.
Following previous works~\cite{chen2024instructzero, hu2024localized, lin2024use}, we set the total query budget to 165, initialize with 40 soft prompts, and evaluate all methods over three different random seeds. 
To ensure a fair comparison, we follow the hyperparameter tuning procedure in~\cite{lin2024use}.
Detailed hyperparameter configurations and experimental settings are provided in the supplement.
% For the score consistency regularization in PRESTO, we set $\gamma_{\text{max}} = 0.1$ and the warm-up duration $T$ to half of the total training epochs across all tasks.
% All methods optimize instruction using the validation set $\mathcal{D}_{\text{val}}$ of each task, and report performance on a held-out test set $\mathcal{D}_{\text{test}}$.
\subsection{Instruction induction results}

\begin{table*}[t]
\caption{Performance on instruction induction tasks. 
% Experimental results for all 30 tasks are in the appendix.
Bolded numbers (blue) indicate the best methods for each task. Scores show the average accuracy with standard error over three runs.}
\centering
\begin{adjustbox}{width=\textwidth}
% \begin{tabular}{lllllll||ll}
\begin{tabular}{lllllll||ll}
\toprule
\textbf{Tasks} & \textbf{APE} & \textbf{InstructZero} & \textbf{INSTINCT} & \textbf{EvoPrompt} & \textbf{ZOPO} & \textbf{OPRO} & \textbf{PRESTO} \\
\midrule
% active\_to\_passive & 98.67 {\scriptsize ± 1.09} & 99.67 {\scriptsize ± 0.27} & 92.00 {\scriptsize ± 6.53} & \gc \textbf{100.00} {\scriptsize ± 0.00} & \gc \textbf{100.00} {\scriptsize ± 0.00} & \gc \textbf{100.00} {\scriptsize ± 0.00} & \gc \textbf{100.00} {\scriptsize ± 0.00} \\
antonyms & 80.67 {\scriptsize ± 0.72} & 75.33 {\scriptsize ± 3.21} & \gc \textbf{83.33} {\scriptsize ± 0.54} & 82.00 {\scriptsize ± 0.47} & 82.67 {\scriptsize ± 1.66} & 80.33 {\scriptsize ± 2.33} & \gc \textbf{83.33} {\scriptsize ± 1.19} \\
auto\_categorization & 26.00 {\scriptsize ± 6.13} & 27.67 {\scriptsize ± 2.60} & 18.67 {\scriptsize ± 0.72} & 29.33 {\scriptsize ± 2.18} & \gc \textbf{31.67} {\scriptsize ± 3.41} & 30.33 {\scriptsize ± 0.72} & \gc \textbf{31.67} {\scriptsize ± 3.41} \\
auto\_debugging & 8.33 {\scriptsize ± 6.80} & 12.50 {\scriptsize ± 5.89} & 10.00 {\scriptsize ± 4.71} & 16.67 {\scriptsize ± 6.80} & 13.33 {\scriptsize ± 7.20} & 8.33 {\scriptsize ± 6.80} & \gc \textbf{20.83} {\scriptsize ± 3.40} \\
cause\_and\_effect & 92.00 {\scriptsize ± 1.89} & 74.67 {\scriptsize ± 4.75} & 76.00 {\scriptsize ± 9.98} & 72.00 {\scriptsize ± 6.80} & 93.33 {\scriptsize ± 2.88} & 38.67 {\scriptsize ± 4.35} & \gc \textbf{94.67} {\scriptsize ± 2.88} \\
common\_concept & 22.36 {\scriptsize ± 2.34} & 15.53 {\scriptsize ± 5.11} & 20.21 {\scriptsize ± 1.19} & 17.99 {\scriptsize ± 6.72} & 21.86 {\scriptsize ± 7.16} & 20.08 {\scriptsize ± 6.70} & \gc \textbf{22.86} {\scriptsize ± 3.27} \\
diff & 18.33 {\scriptsize ± 6.87} & 53.00 {\scriptsize ± 20.37} & 81.67 {\scriptsize ± 13.76} & 7.00 {\scriptsize ± 5.72} & 88.33 {\scriptsize ± 5.93} & 64.33 {\scriptsize ± 23.91} & \gc \textbf{98.00} {\scriptsize ± 0.82} \\
% first\_word\_letter & 99.33 {\scriptsize ± 0.54} & \gc \textbf{100.00} {\scriptsize ± 0.00} & \gc \textbf{100.00} {\scriptsize ± 0.00} & \gc \textbf{100.00} {\scriptsize ± 0.00} & 95.33 {\scriptsize ± 1.96} & \gc \textbf{100.00} {\scriptsize ± 0.00} & \gc \textbf{100.00} {\scriptsize ± 0.00} \\
informal\_to\_formal & 57.59 {\scriptsize ± 2.40} & 51.53 {\scriptsize ± 4.62} & 48.93 {\scriptsize ± 3.46} & 42.87 {\scriptsize ± 2.03} & \gc \textbf{58.93} {\scriptsize ± 4.83} & 50.02 {\scriptsize ± 2.63} & 52.77 {\scriptsize ± 5.46} \\
% larger\_animal & \gc \textbf{93.33} {\scriptsize ± 0.98} & 73.33 {\scriptsize ± 11.06} & 76.00 {\scriptsize ± 6.94} & 49.33 {\scriptsize ± 2.84} & 79.33 {\scriptsize ± 9.27} & 84.67 {\scriptsize ± 0.72} & 79.67 {\scriptsize ± 9.30} \\
letters\_list & 99.00 {\scriptsize ± 0.82} & 99.00 {\scriptsize ± 0.47} & 97.67 {\scriptsize ± 1.52} & 73.67 {\scriptsize ± 9.69} & 98.67 {\scriptsize ± 1.09} & 99.00 {\scriptsize ± 0.47} & \gc \textbf{99.33} {\scriptsize ± 0.54} \\
negation & 83.33 {\scriptsize ± 1.19} & 81.67 {\scriptsize ± 3.95} & 76.67 {\scriptsize ± 4.77} & 71.67 {\scriptsize ± 1.19} & 77.33 {\scriptsize ± 4.63} & 73.33 {\scriptsize ± 4.23} & \gc \textbf{84.00} {\scriptsize ± 2.16} \\
% num\_to\_verbal & 96.33 {\scriptsize ± 2.60} & 99.33 {\scriptsize ± 0.27} & \gc \textbf{100.00} {\scriptsize ± 0.00} & \gc \textbf{100.00} {\scriptsize ± 0.00} & \gc \textbf{100.00} {\scriptsize ± 0.00} & 99.67 {\scriptsize ± 0.27} & \gc \textbf{100.00} {\scriptsize ± 0.00} \\
object\_counting & 37.33 {\scriptsize ± 5.50} & 46.00 {\scriptsize ± 5.72} & \gc \textbf{48.67} {\scriptsize ± 3.21} & 28.67 {\scriptsize ± 2.23} & 34.00 {\scriptsize ± 4.08} & 31.00 {\scriptsize ± 3.86} & 45.67 {\scriptsize ± 4.38} \\
odd\_one\_out & 51.33 {\scriptsize ± 14.43} & 46.67 {\scriptsize ± 5.76} & 60.00 {\scriptsize ± 7.12} & 68.00 {\scriptsize ± 1.89} & 58.67 {\scriptsize ± 7.14} & 47.33 {\scriptsize ± 10.39} & \gc \textbf{70.00} {\scriptsize ± 0.94} \\
orthography\_starts\_with & 46.00 {\scriptsize ± 8.18} & 35.00 {\scriptsize ± 3.56} & 54.67 {\scriptsize ± 8.20} & 42.00 {\scriptsize ± 15.28} & 54.67 {\scriptsize ± 3.66} & 22.33 {\scriptsize ± 10.18} & \gc \textbf{57.33} {\scriptsize ± 6.08} \\
% periodic\_elements & 99.33 {\scriptsize ± 0.54} & 93.33 {\scriptsize ± 3.03} & 98.67 {\scriptsize ± 1.09} & 70.67 {\scriptsize ± 19.14} & \gc \textbf{100.00} {\scriptsize ± 0.00} & 99.33 {\scriptsize ± 0.54} & 99.33 {\scriptsize ± 0.54} \\
rhymes & 69.33 {\scriptsize ± 16.41} & 81.67 {\scriptsize ± 10.69} & \gc \textbf{98.67} {\scriptsize ± 0.72} & 93.67 {\scriptsize ± 1.96} & 83.33 {\scriptsize ± 6.87} & 77.00 {\scriptsize ± 15.25} & 85.00 {\scriptsize ± 7.41} \\
second\_word\_letter & 72.67 {\scriptsize ± 10.88} & 40.67 {\scriptsize ± 5.99} & 48.00 {\scriptsize ± 22.38} & 33.00 {\scriptsize ± 7.93} & 68.00 {\scriptsize ± 17.75} & 22.00 {\scriptsize ± 14.73} & \gc \textbf{77.00} {\scriptsize ± 12.57} \\
sentence\_similarity & \gc \textbf{29.00} {\scriptsize ± 5.44} & 17.33 {\scriptsize ± 4.75} & 11.33 {\scriptsize ± 5.42} & \gc \textbf{29.00} {\scriptsize ± 0.47} & 4.33 {\scriptsize ± 3.54} & 6.67 {\scriptsize ± 5.44} & 21.67 {\scriptsize ± 8.49} \\
% sentiment & 88.00 {\scriptsize ± 0.47} & 90.67 {\scriptsize ± 0.98} & 90.33 {\scriptsize ± 1.36} & 87.67 {\scriptsize ± 0.72} & \gc \textbf{91.00} {\scriptsize ± 0.47} & 89.33 {\scriptsize ± 2.18} & \gc \textbf{91.00} {\scriptsize ± 0.00} \\
% singular\_to\_plural & 99.33 {\scriptsize ± 0.54} & 96.67 {\scriptsize ± 1.91} & 98.33 {\scriptsize ± 0.72} & \gc \textbf{100.00} {\scriptsize ± 0.00} & 99.33 {\scriptsize ± 0.27} & 91.67 {\scriptsize ± 4.01} & \gc \textbf{100.00} {\scriptsize ± 0.00} \\
sum & 24.00 {\scriptsize ± 14.61} & 55.00 {\scriptsize ± 23.92} & 99.33 {\scriptsize ± 0.54} & 66.67 {\scriptsize ± 27.22} & \gc \textbf{100.00} {\scriptsize ± 0.00} & 91.33 {\scriptsize ± 3.78} & 94.67 {\scriptsize ± 4.35} \\
synonyms & 10.00 {\scriptsize ± 4.50} & 22.67 {\scriptsize ± 5.62} & 25.00 {\scriptsize ± 8.83} & \gc \textbf{25.33} {\scriptsize ± 7.98} & 24.33 {\scriptsize ± 2.76} & 12.67 {\scriptsize ± 0.72} & 18.33 {\scriptsize ± 1.91} \\
taxonomy\_animal & 43.67 {\scriptsize ± 15.96} & 44.33 {\scriptsize ± 17.72} & 92.00 {\scriptsize ± 3.77} & 34.00 {\scriptsize ± 15.08} & 69.00 {\scriptsize ± 24.10} & 73.67 {\scriptsize ± 8.09} & \gc \textbf{99.67} {\scriptsize ± 0.27} \\
% translation\_en-de & 84.67 {\scriptsize ± 1.19} & 74.00 {\scriptsize ± 3.30} & 85.33 {\scriptsize ± 0.72} & 77.33 {\scriptsize ± 2.60} & 83.67 {\scriptsize ± 1.19} & 57.00 {\scriptsize ± 20.82} & \gc \textbf{85.67} {\scriptsize ± 0.54} \\
% translation\_en-es & \gc \textbf{90.67} {\scriptsize ± 0.98} & 83.33 {\scriptsize ± 3.07} & 88.33 {\scriptsize ± 1.78} & 83.67 {\scriptsize ± 2.76} & 89.00 {\scriptsize ± 0.47} & 85.33 {\scriptsize ± 0.27} & 86.00 {\scriptsize ± 2.05} \\
% translation\_en-fr & 87.33 {\scriptsize ± 0.72} & 82.00 {\scriptsize ± 0.94} & \gc \textbf{88.00} {\scriptsize ± 1.63} & 84.00 {\scriptsize ± 2.05} & 87.67 {\scriptsize ± 1.91} & 84.67 {\scriptsize ± 3.14} & 83.00 {\scriptsize ± 2.36} \\
word\_sorting & 54.00 {\scriptsize ± 15.41} & 39.67 {\scriptsize ± 12.11} & 27.33 {\scriptsize ± 7.37} & \gc \textbf{71.00} {\scriptsize ± 4.50} & 54.00 {\scriptsize ± 15.06} & 36.33 {\scriptsize ± 11.49} & 53.33 {\scriptsize ± 8.38} \\
word\_unscrambling & 28.00 {\scriptsize ± 4.78} & 38.00 {\scriptsize ± 3.74} & 42.33 {\scriptsize ± 8.59} & 23.00 {\scriptsize ± 9.57} & \gc \textbf{52.00} {\scriptsize ± 7.79} & 43.00 {\scriptsize ± 1.25} & 48.00 {\scriptsize ± 7.59} \\
\midrule
\textbf{\# best-performing tasks} & 1 & 0 & 3 & 3 & 4 & 0 & \gc \textbf{12} \\
\textbf{Average Rank} & 4.25 & 4.80 & 3.70 & 4.70 & 3.05 & 5.20 & \gc \textbf{1.90} \\
\bottomrule
\end{tabular}
\end{adjustbox}
\label{tab:main results}
\vspace{-0.05em}
\end{table*}
Here we provide the results of our proposed method, PRESTO, compared with six strong baselines on instruction induction tasks.
To enhance readability, we report results on a subset of 20 following previous works~\cite{lin2024use, hu2024localized}.
The full results for all 30 tasks are provided in the appendix.
Table~\ref{tab:main results} shows that PRESTO achieves the highest accuracy on 12 out of the 20 tasks, which is three times more than the second-best method, ZOPO.
In addition, PRESTO attains the best average rank of 1.90, outperforming all baselines by a clear margin; the next best, ZOPO, has an average rank of 3.05, followed by INSTINCT at 3.70.
These results highlight the strong performance of PRESTO on individual tasks and its robustness across a wide range of instruction induction tasks.
In the full set of 30 tasks, PRESTO also consistently outperforms other baselines with a large margin in the number of best-performing tasks and average rank.
\subsection{Chain-of-Thought Prompting Results}

\begin{table}[t]
\centering
\caption{
Performance of different CoT prompts on three math reasoning datasets. 
% We report the accuracy achieved when using each instruction. 
The best result for each dataset is in \textbf{bold}, and the second best is \underline{underlined}. 
% If multiple methods achieve the same score, all can be shown in bold or underlined.
}
\label{tab:cot_prompts}
\begin{adjustbox}{width=0.82\textwidth}
\begin{tabular}{l|l|p{7.5cm}|c}
\toprule
\textbf{Method} & \textbf{Dataset} & \textbf{Best instruction} & \textbf{Accuracy} \\
\midrule
Hand-crafted & GSM8K & Let’s think step by step & 0.9121 \\
InstructZero & GSM8K & Let’s think step by step to solve the math problem & 0.9083 \\
INSTINCT & GSM8K & Let’s break down and solve the problem & 0.9098 \\
ZOPO & GSM8K & Let's break it down and find the solution & \textbf{0.9143} \\
PRESTO (Ours) & GSM8K & Let's break it down together & \underline{0.9128} \\
\midrule
Hand-crafted & AQUA-RAT & Let’s think step by step. & 0.7402 \\
InstructZero & AQUA-RAT & Let’s break it down and find the solution & 0.7480 \\
INSTINCT & AQUA-RAT & Let's break it down step by step. I am ready to solve the problem. & 0.7480 \\
ZOPO & AQUA-RAT & Let's break it down mathematically. & \underline{0.7520} \\
PRESTO (Ours) & AQUA-RAT & Let's solve it together. & \textbf{0.7756} \\
\midrule
Hand-crafted & SVAMP & Let’s think step by step. & \underline{0.9375} \\
InstructZero & SVAMP & Let's crack the code! & \textbf{0.9400} \\
INSTINCT & SVAMP & Let's break it down step by step & \underline{0.9375} \\
ZOPO & SVAMP & I see what you're doing there & \textbf{0.9400} \\
PRESTO (Ours) & SVAMP & Let’s use the formula & \textbf{0.9400} \\
\bottomrule
\end{tabular}
\end{adjustbox}
\vspace{-0.7em}
\end{table}
We evaluate the quality of the optimized instructions by measuring their effectiveness as chain-of-thought (CoT)~\cite{wei2022chain} prompts on three math reasoning benchmarks: GSM8K~\cite{cobbe2021training}, AQUA-RAT~\cite{ling2017program}, and SVAMP~\cite{patel2021nlp}. 
We compare our method with three baselines that use soft prompts (InstructZero~\cite{chen2024instructzero}, INSTINCT~\cite{lin2024use}, and ZOPO~\cite{hu2024localized}), as well as a standard hand-crafted prompt~\cite{kojima2022large}.
Table~\ref{tab:cot_prompts} demonstrates that our PRESTO outperforms or matches the best-performing baselines across all datasets.
In particular, it achieves the highest accuracy on AQUA-RAT (0.7756) and ties for the best result on SVAMP (0.9400), while remaining competitive on GSM8K. 
These results indicate that the instructions optimized by our method are also effective when used as CoT prompts.
\section{Analysis}
\subsection{Ablation Study}
\definecolor{mygreen}{RGB}{0,100,0}
\begin{table}[t]
\centering
\caption{Ablation study of \textbf{PRESTO}. We incrementally add score sharing (\textbf{SS}, Sec.~\ref{main_sec:ScoreAssignment}), preimage-based initialization (\textbf{Init}, Sec.~\ref{main_sec:GroupInitialization}), and consistency regularization (\textbf{Reg}, Sec.~\ref{main_sec:NeuralUCB}) to a vanilla baseline.
}
\label{tab:ablation}
\begin{adjustbox}{width=0.8\textwidth}
\begin{tabular}{l|ccc|ccc}
\toprule
% \textbf{Model} & \textbf{SS} (Sec.~\ref{main_sec:ScoreAssignment}) & \textbf{Init} (Sec.~\ref{main_sec:GroupInitialization}) & \textbf{Reg} (Sec.~\ref{main_sec:NeuralUCB}) & \# Wins & Avg. Rank & Avg. acc. \\
\textbf{Model} & \textbf{     SS     } & \textbf{     Reg     } & \textbf{     Init     } & \# Wins & Avg. Rank & Avg. acc. \\
\midrule
Vanilla                 & \textcolor{red}{\ding{55}} & \textcolor{red}{\ding{55}} & \textcolor{red}{\ding{55}} & 0 & 4.55 & 51.91 \\
+ SS          & \textcolor{mygreen}{\ding{51}} & \textcolor{red}{\ding{55}} & \textcolor{red}{\ding{55}} & 3 & 3.10 & 59.57\\
+ SS + Reg        & \textcolor{mygreen}{\ding{51}} & \textcolor{mygreen}{\ding{51}} & \textcolor{red}{\ding{55}} & 4 & 2.65 & 61.77\\
+ SS + Init        & \textcolor{mygreen}{\ding{51}} & \textcolor{red}{\ding{55}} & \textcolor{mygreen}{\ding{51}} & 4 & 2.30 & 61.82\\
+ SS + Init + Reg (\textbf{Ours})   & \textcolor{mygreen}{\ding{51}} & \textcolor{mygreen}{\ding{51}} & \textcolor{mygreen}{\ding{51}} & \textbf{9} & \textbf{2.20} & \textbf{62.91}\\
\bottomrule
\end{tabular}
\end{adjustbox}
\vspace{-0.7em}
\end{table}
We perform an ablation study to analyze the contribution of each component in our method over the 20 instruction induction tasks used in Table~\ref{tab:main results} over 3 random seeds.
Starting from a vanilla baseline without our techniques, we incrementally add: (1) score sharing method (Section~\ref{main_sec:ScoreAssignment}), (2) preimage-based initialization (Section~\ref{main_sec:GroupInitialization}), and (3) score consistency regularization (Section~\ref{main_sec:NeuralUCB}). 
The full model with all components combined corresponds to our proposed method, PRESTO. 
As shown in Table~\ref{tab:ablation}, each component contributes to performance improvement. 
In particular, introducing score sharing significantly boosts accuracy from 51.91 to 59.57 (+7.66) and improves average rank from 4.55 to 3.10 (-1.45), indicating its strong impact.
Our PRESTO achieves the best results overall, with the highest number of wins and the lowest average rank across tasks.
\subsection{Impact of score sharing method}
% \begin{wrapfigure}{rt}{0.48\textwidth}
%     \centering
%     \vspace{-14pt}
%     \includegraphics[width=0.48\textwidth]{Figures/Num_observed.png}
%     \caption{
%         Average number of soft prompts with assigned scores after optimization, averaged across tasks.
%     } 
%     \label{fig:observed_score}
%     \vspace{-14pt}
% \end{wrapfigure}

% \begin{figure}[t]
%     % \centering
%     \includegraphics[width=0.3\textwidth]{Figures/Num_observed.png}
%     \caption{
%         Average number of soft prompts with assigned scores after optimization, averaged across tasks.
%     } 
%     \label{fig:observed_score}
% \end{figure}

% \begin{wrapfigure}{rt}{0.48\textwidth}
%     \centering
%     \vspace{-14pt}
%     \includegraphics[width=0.48\textwidth]{Figures/prediction.png}
%     \caption{
%         Performance of score predictor trained with diverse methods. We measure the prediction performance using RMSE, the lower is better.
%     } 
%     \label{fig:score_prediction}
%     \vspace{-14pt}
% \end{wrapfigure}

% \begin{minipage}[t]{0.3\textwidth}
%     % \begin{figure}[H]
%         \includegraphics[width=\linewidth]{Figures/prediction.png}
%         \captionof{figure}{
%         Performance of score predictor trained with diverse methods. We measure the prediction performance using RMSE, the lower is better.
%         }
%         \label{fig:score_prediction}
%     % \end{figure}
% \end{minipage}

\begin{figure}[t]
\begin{minipage}[h]{0.47\textwidth}
    \centering
    \vspace{0.8em}
    \includegraphics[width=0.70\linewidth]{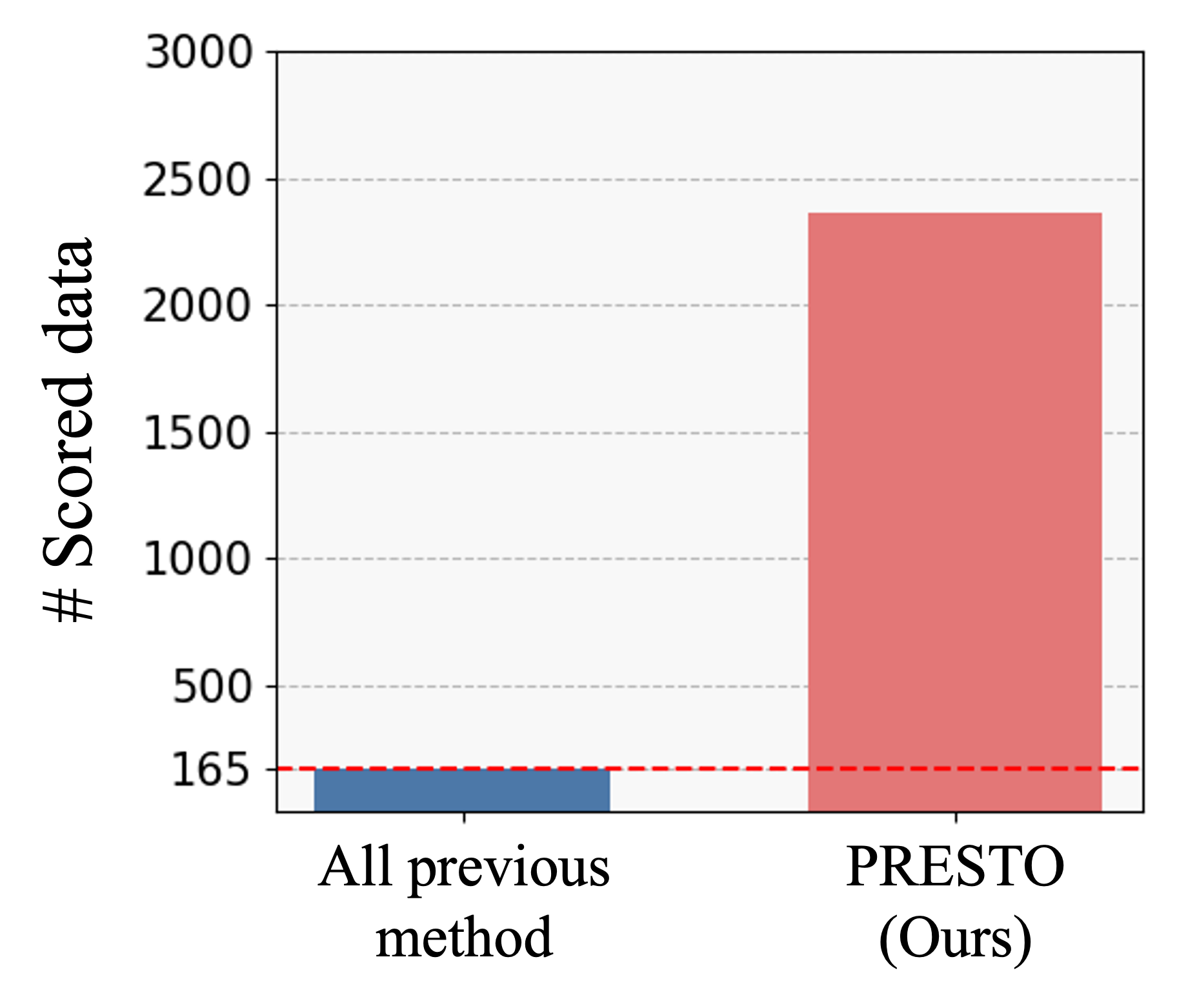}
    \vspace{-0.5em}
    \captionof{figure}{
        Average number of scored soft prompts after optimization across all tasks.
    }
    \label{fig:observed_score}
    \end{minipage}
\hfill
\begin{minipage}[h]{0.5\textwidth}
    \centering
    \hspace{-3.0em}
    \includegraphics[width=0.64\linewidth]{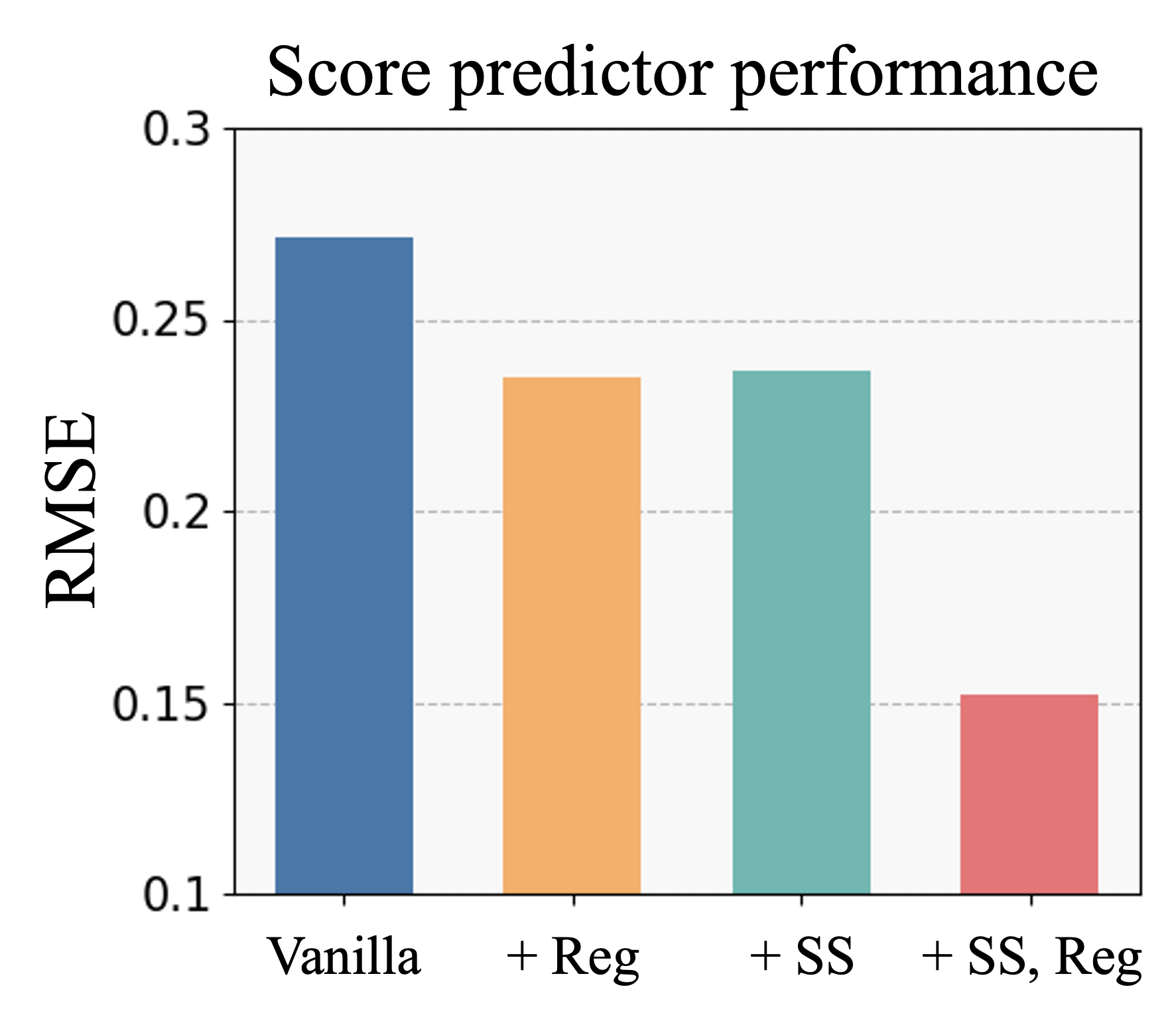}
    \vspace{0.6em}
    \captionof{figure}{
    Performance of score predictor trained with diverse methods.
    }
    \label{fig:score_prediction}
\end{minipage}
\vspace{-0.8em}
\end{figure}
We report the average number of soft prompts with assigned scores after the optimization process, comparing our method with baselines across all 30 tasks. 
The reported count includes soft prompts that were scored either directly through black-box evaluation or indirectly via score sharing.
% As shown in Figure~\ref{fig:observed_score}, our method assigns scores to over 2,300 soft prompts on average, whereas previous methods yield only 165 scored data points—the same as the query budget in our setting.
As shown in Figure~\ref{fig:observed_score}, our method assigns scores to over 2,300 soft prompts on average, 14× more than previous methods, which yield only 165 scored data points, equal to the query budget in our setting.
The large amount of scored data enables the score predictor to learn the objective function more effectively, which in turn facilitates more successful optimization.
This analysis demonstrates that our score-sharing method can significantly increase the amount of scored data without requiring additional black-box queries.
\subsection{Visualization of Preimage-Based Initialization}
\label{analysis:Preimage_initialization}
\begin{figure}[t]
    \centering
    \includegraphics[width=0.9\textwidth]{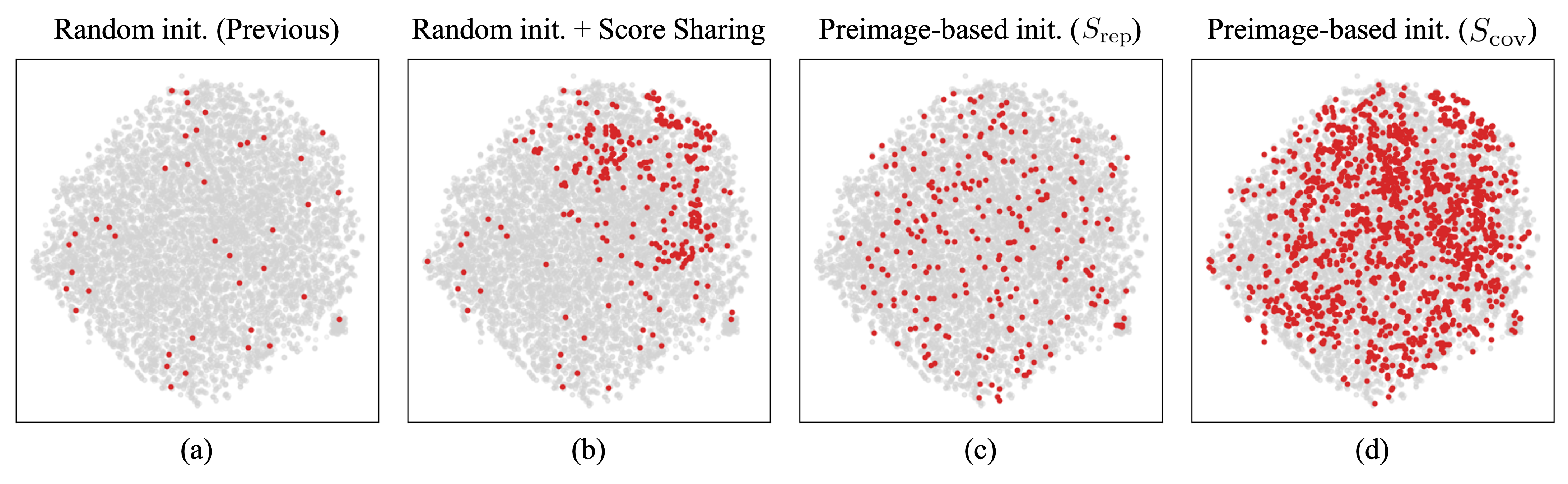}
    \caption{
    Visualization of the initial data distribution under different initialization. We plot the entire soft prompt embedding candidate set $G^\text{total}$ using t-SNE, and highlight the selected initial data in red. %From left to right: (a) random initialization, (b)  apply score sharing to (a), (c) Preimage-based initialization using $S_{\text{rep}}$, and (d) Preimage-based initialization using $S_{\text{cov}}$. 
    } 
    \label{fig:group_initialization}
    \vspace{-1.3em}
\end{figure}
We present a qualitative analysis of how score sharing and preimage-based initialization influence the distribution of initial soft prompts.
    Figure~\ref{fig:group_initialization} visualizes the distribution of initial soft prompts under four settings: (1) random initialization, (2) random initialization with score sharing (Section~\ref{main_sec:ScoreAssignment}), (3) preimage-based initialization using $S_{\text{rep}}$ only, and (4) $S_{\text{cov}} = S_{\text{rep}} + S_{\text{size}}$ (Section~\ref{main_sec:GroupInitialization}) in "objective counting" task. 
To visualize the spatial distribution of soft prompt embeddings, we employ t-SNE.
Compared to random initialization in prior works, score sharing enlarges the size of the initial dataset without additional black-box queries.
Furthermore, selecting initial data using $S_{\text{rep}}$ leads to better coverage of the soft prompt space than naive score sharing. 
Finally, our proposed preimage-based initialization method that utilizes $S_{\text{cov}}$ achieves the densest and comprehensive coverage of the search space.
It shows that our preimage-based initialization method effectively selects the initial data points that densely and evenly cover the search space.
\subsection{Score Predictor Performance Enhancement}
To analyze how score sharing (Section~\ref{main_sec:ScoreAssignment}) and score consistency regularization (Section~\ref{main_sec:NeuralUCB}) influence the quality of the score predictor, we evaluate its prediction performance under different training configurations. 
Figure~\ref{fig:score_prediction} reports the root mean squared error (RMSE), where lower values indicate higher prediction accuracy.
We use 100 randomly selected soft prompts as training data and another 100 as test data for the objective counting task.
As shown in Figure~\ref{fig:score_prediction}, applying either score sharing or score consistency regularization improves the score predictor’s performance, reducing the RMSE from approximately 0.27 (vanilla) to around 0.23. 
When both techniques are applied together, the RMSE further decreases to approximately 0.15, indicating a strong complementary effect.
The results demonstrate that expanding the training set without requiring additional black-box queries through score sharing and incorporating the preimage structure as a prior via score consistency regularization are both crucial for enhancing the score predictor’s performance.

\section{Conclusion}
\label{sec:conclusion}
We propose PRESTO, a preimage-informed instruction optimization framework that explicitly leverages this many-to-one structure via preimage. 
% Our PRESTO consists of three key components. 
% First, we propose a preimage-based score sharing method that shares the observed score to all other members in a preimage, which significantly increases the amount of scored data without requiring additional black-box queries.
% Second, we present the preimage-based initialization method, selecting initial data to maximize search space coverage by leveraging the preimage structure.
% Finally, we introduce a consistency regularization term that encourages consistent predictions within unobserved groups, improving the performance and generalization of the score predictor.
PRESTO consists of three components that leverage the preimage structure: score sharing to propagate labels within each preimage, preimage-based initialization to improve search space coverage, and consistency regularization to align predictions within unscored preimages. 
PRESTO achieves state-of-the-art performance on 33 instruction optimization tasks, and our comprehensive analysis supports its effectiveness and robustness.

% \section{Limitations and broader impacts}
\textbf{Limitations and broader impacts}

Our method introduces preimage-based score sharing to enlarge the number of data, which incurs mild computational overhead compared to simpler baselines. Moreover, its benefits are more pronounced when applied to a large candidate set, as score sharing is most effective when many soft prompts map to the same instruction.

In terms of broader impact, this work aims to make black-box LLM optimization more data-efficient, which can reduce the cost of experimentation and improve accessibility for researchers with limited resources. However, as with any optimization technique for LLMs, there is a risk that improved performance could be applied in ways that reinforce biases or generate harmful content. Careful deployment and alignment with responsible AI principles are necessary.

\subsubsection*{Acknowledgement}
This research was supported by the ASTRA Project through the National Research Foundation (NRF) funded by the Ministry of Science and ICT (No. RS-2024-00439619).

\bibliographystyle{unsrt} % Sorting References
\bibliography{Reference}

\newpage
\clearpage
% \begin{document}
% \maketitle
% \paragraph{Summary.}
% In this supplement, we present the experimental details, additional results, and analysis:
% (A) Experimental details and additional results, (B) details of NeuralUCB, (C) full experimental results of the ablation study (D) efficiency analysis of each component in PRESTO (E) impact of hyperparameters to the preimage structure (F) preimage structure in different white-box LLMs (G) experimental results on different combinations of white-box LLMs and black-box LLMs (H) impact of hyperparameters in score consiscency regularization (I) details of preimage-based initialization method (J) best instructions for each task discovered by PRESTO and (K) instruction generation format.
\section{Experimental Details and Additional Results}
\label{supp_sec:experimental}
\noindent\textbf{Experimental Settings}
% Hyperparameters / Score metric / black box GPT 4.1로 쓴 이유
Following previous works~\cite{chen2024instructzero,lin2024use, hu2024localized}, we select $N$ soft prompts by first sampling $N$ vectors from a scrambled Sobol sequence in a low-dimensional space, and then mapping them to the soft prompt space using a fixed random projection matrix. 
Since the dimensionality of this low-dimensional space—$\textit{i.e.,}$ the intrinsic dimension—has been shown to play a critical role in optimization performance, we follow previous studies and perform a grid search over both the intrinsic dimension and the number of soft prompt tokens. 
Specifically, we search over intrinsic dimensions of {10, 50, 100} and soft prompt token counts of {3, 5, 10} for the instruction induction tasks, and we fix the intrinsic dimension to 1000 in Chain-of-Thought tasks following previous works. 
We select the hyperparameters based on validation performance from the first random seed and apply the same hyperparameters to the remaining two seeds. As in prior work, we fix $N$=10,000.
For dataset split, we follow previous works~\cite{lin2024use}.
All experiments were conducted on an NVIDIA A6000 GPU.
Our implementation is built upon the codebase of INSTINCT~\cite{lin2024use}.

\noindent\textbf{Evaluation Metrics}
We use the F1 score for common\_concept, informal\_to\_formal. 
For orthography\_starts\_with and taxonomy\_animal, we use exact set matching. 
For synonyms, we evaluate whether the output label is contained in the model's prediction. 
For all remaining instruction induction tasks, we adopt the exact match metric.
For Chain-of-Thought tasks, we extract the final answer using the GPT-4.1 and use exact matching to measure the accuracy.

\noindent\textbf{Details of Baselines}
In the instruction induction tasks, we compare our PRESTO with six strong instruction optimization methods.
$\textbf{APE}$~\cite{pryzant2023automatic} generates instructions by leveraging predefined templates and augmented exemplars, and selects high-performing instructions from LLM-proposed candidates.
$\textbf{InstructZero}$~\cite{chen2024instructzero} takes a Bayesian Optimization, aiming to generate optimal instructions for black-box LLM by optimizing the soft prompt, which is taken as input for the white-box LLM. 
$\textbf{INSTINCT}$~\cite{lin2024use} leverages NeuralUCB to optimize the soft prompts, while taking the white-box LLM as a feature extractor for score prediction.
$\textbf{EvoPrompt}$~\cite{guo2024connecting} explores a population of prompt candidates using evolutionary algorithms to identify high-performing prompts.
$\textbf{ZOPO}$~\cite{hu2024localized} employs a Neural Tangent Kernel-guided Gaussian process to efficiently search for locally optimal soft prompts.
Finally, $\textbf{OPRO}$~\cite{yang2023large} iteratively updates the optimization trajectory and exemplars within the meta-prompt during the optimization, enabling the LLM to progressively refine its search.

\noindent\textbf{Experimental results for all 30 tasks}
\begin{table*}[ht]
\caption{Performance on 30 instruction induction tasks. 
Bolded numbers with blue colors indicate the best algorithm for each task. Scores show the average accuracy with standard error over three runs.}
\centering
\begin{adjustbox}{width=\textwidth}
\begin{tabular}{lllllll||ll}
\toprule
\textbf{Tasks} & \textbf{APE} & \textbf{InstructZero} & \textbf{INSTINCT} & \textbf{EvoPrompt} & \textbf{ZOPO} & \textbf{OPRO} & \textbf{PRESTO} \\
\midrule
active\_to\_passive & 98.67 {\scriptsize ± 1.09} & 99.67 {\scriptsize ± 0.27} & 92.00 {\scriptsize ± 6.53} & \gc \textbf{100.00} {\scriptsize ± 0.00} & \gc \textbf{100.00} {\scriptsize ± 0.00} & \gc \textbf{100.00} {\scriptsize ± 0.00} & \gc \textbf{100.00} {\scriptsize ± 0.00} \\
antonyms & 80.67 {\scriptsize ± 0.72} & 75.33 {\scriptsize ± 3.21} & \gc \textbf{83.33} {\scriptsize ± 0.54} & 82.00 {\scriptsize ± 0.47} & 82.67 {\scriptsize ± 1.66} & 80.33 {\scriptsize ± 2.33} & \gc \textbf{83.33} {\scriptsize ± 1.19} \\
auto\_categorization & 26.00 {\scriptsize ± 6.13} & 27.67 {\scriptsize ± 2.60} & 18.67 {\scriptsize ± 0.72} & 29.33 {\scriptsize ± 2.18} & \gc \textbf{31.67} {\scriptsize ± 3.41} & 30.33 {\scriptsize ± 0.72} & \gc \textbf{31.67} {\scriptsize ± 3.41} \\
auto\_debugging & 8.33 {\scriptsize ± 6.80} & 12.50 {\scriptsize ± 5.89} & 10.00 {\scriptsize ± 4.71} & 16.67 {\scriptsize ± 6.80} & 13.33 {\scriptsize ± 7.20} & 8.33 {\scriptsize ± 6.80} & \gc \textbf{20.83} {\scriptsize ± 3.40} \\
cause\_and\_effect & 92.00 {\scriptsize ± 1.89} & 74.67 {\scriptsize ± 4.75} & 76.00 {\scriptsize ± 9.98} & 72.00 {\scriptsize ± 6.80} & 93.33 {\scriptsize ± 2.88} & 38.67 {\scriptsize ± 4.35} & \gc \textbf{94.67} {\scriptsize ± 2.88} \\
common\_concept & 22.36 {\scriptsize ± 2.34} & 15.53 {\scriptsize ± 5.11} & 20.21 {\scriptsize ± 1.19} & 17.99 {\scriptsize ± 6.72} & 21.86 {\scriptsize ± 7.16} & 20.08 {\scriptsize ± 6.70} & \gc \textbf{22.86} {\scriptsize ± 3.27} \\
diff & 18.33 {\scriptsize ± 6.87} & 53.00 {\scriptsize ± 20.37} & 81.67 {\scriptsize ± 13.76} & 7.00 {\scriptsize ± 5.72} & 88.33 {\scriptsize ± 5.93} & 64.33 {\scriptsize ± 23.91} & \gc \textbf{98.00} {\scriptsize ± 0.82} \\
first\_word\_letter & 99.33 {\scriptsize ± 0.54} & \gc \textbf{100.00} {\scriptsize ± 0.00} & \gc \textbf{100.00} {\scriptsize ± 0.00} & \gc \textbf{100.00} {\scriptsize ± 0.00} & 95.33 {\scriptsize ± 1.96} & \gc \textbf{100.00} {\scriptsize ± 0.00} & \gc \textbf{100.00} {\scriptsize ± 0.00} \\
informal\_to\_formal & 57.59 {\scriptsize ± 2.40} & 51.53 {\scriptsize ± 4.62} & 48.93 {\scriptsize ± 3.46} & 42.87 {\scriptsize ± 2.03} & \gc \textbf{58.93} {\scriptsize ± 4.83} & 50.02 {\scriptsize ± 2.63} & 52.77 {\scriptsize ± 5.46} \\
larger\_animal & \gc \textbf{93.33} {\scriptsize ± 0.98} & 73.33 {\scriptsize ± 11.06} & 76.00 {\scriptsize ± 6.94} & 49.33 {\scriptsize ± 2.84} & 79.33 {\scriptsize ± 9.27} & 84.67 {\scriptsize ± 0.72} & 79.67 {\scriptsize ± 9.30} \\
letters\_list & 99.00 {\scriptsize ± 0.82} & 99.00 {\scriptsize ± 0.47} & 97.67 {\scriptsize ± 1.52} & 73.67 {\scriptsize ± 9.69} & 98.67 {\scriptsize ± 1.09} & 99.00 {\scriptsize ± 0.47} & \gc \textbf{99.33} {\scriptsize ± 0.54} \\
negation & 83.33 {\scriptsize ± 1.19} & 81.67 {\scriptsize ± 3.95} & 76.67 {\scriptsize ± 4.77} & 71.67 {\scriptsize ± 1.19} & 77.33 {\scriptsize ± 4.63} & 73.33 {\scriptsize ± 4.23} & \gc \textbf{84.00} {\scriptsize ± 2.16} \\
num\_to\_verbal & 96.33 {\scriptsize ± 2.60} & 99.33 {\scriptsize ± 0.27} & \gc \textbf{100.00} {\scriptsize ± 0.00} & \gc \textbf{100.00} {\scriptsize ± 0.00} & \gc \textbf{100.00} {\scriptsize ± 0.00} & 99.67 {\scriptsize ± 0.27} & \gc \textbf{100.00} {\scriptsize ± 0.00} \\
object\_counting & 37.33 {\scriptsize ± 5.50} & 46.00 {\scriptsize ± 5.72} & \gc \textbf{48.67} {\scriptsize ± 3.21} & 28.67 {\scriptsize ± 2.23} & 34.00 {\scriptsize ± 4.08} & 31.00 {\scriptsize ± 3.86} & 45.67 {\scriptsize ± 4.38} \\
odd\_one\_out & 51.33 {\scriptsize ± 14.43} & 46.67 {\scriptsize ± 5.76} & 60.00 {\scriptsize ± 7.12} & 68.00 {\scriptsize ± 1.89} & 58.67 {\scriptsize ± 7.14} & 47.33 {\scriptsize ± 10.39} & \gc \textbf{70.00} {\scriptsize ± 0.94} \\
orthography\_starts\_with & 46.00 {\scriptsize ± 8.18} & 35.00 {\scriptsize ± 3.56} & 54.67 {\scriptsize ± 8.20} & 42.00 {\scriptsize ± 15.28} & 54.67 {\scriptsize ± 3.66} & 22.33 {\scriptsize ± 10.18} & \gc \textbf{57.33} {\scriptsize ± 6.08} \\
periodic\_elements & 99.33 {\scriptsize ± 0.54} & 93.33 {\scriptsize ± 3.03} & 98.67 {\scriptsize ± 1.09} & 70.67 {\scriptsize ± 19.14} & \gc \textbf{100.00} {\scriptsize ± 0.00} & 99.33 {\scriptsize ± 0.54} & 99.33 {\scriptsize ± 0.54} \\
rhymes & 69.33 {\scriptsize ± 16.41} & 81.67 {\scriptsize ± 10.69} & \gc \textbf{98.67} {\scriptsize ± 0.72} & 93.67 {\scriptsize ± 1.96} & 83.33 {\scriptsize ± 6.87} & 77.00 {\scriptsize ± 15.25} & 85.00 {\scriptsize ± 7.41} \\
second\_word\_letter & 72.67 {\scriptsize ± 10.88} & 40.67 {\scriptsize ± 5.99} & 48.00 {\scriptsize ± 22.38} & 33.00 {\scriptsize ± 7.93} & 68.00 {\scriptsize ± 17.75} & 22.00 {\scriptsize ± 14.73} & \gc \textbf{77.00} {\scriptsize ± 12.57} \\
sentence\_similarity & \gc \textbf{29.00} {\scriptsize ± 5.44} & 17.33 {\scriptsize ± 4.75} & 11.33 {\scriptsize ± 5.42} & \gc \textbf{29.00} {\scriptsize ± 0.47} & 4.33 {\scriptsize ± 3.54} & 6.67 {\scriptsize ± 5.44} & 21.67 {\scriptsize ± 8.49} \\
sentiment & 88.00 {\scriptsize ± 0.47} & 90.67 {\scriptsize ± 0.98} & 90.33 {\scriptsize ± 1.36} & 87.67 {\scriptsize ± 0.72} & \gc \textbf{91.00} {\scriptsize ± 0.47} & 89.33 {\scriptsize ± 2.18} & \gc \textbf{91.00} {\scriptsize ± 0.00} \\
singular\_to\_plural & 99.33 {\scriptsize ± 0.54} & 96.67 {\scriptsize ± 1.91} & 98.33 {\scriptsize ± 0.72} & \gc \textbf{100.00} {\scriptsize ± 0.00} & 99.33 {\scriptsize ± 0.27} & 91.67 {\scriptsize ± 4.01} & \gc \textbf{100.00} {\scriptsize ± 0.00} \\
sum & 24.00 {\scriptsize ± 14.61} & 55.00 {\scriptsize ± 23.92} & 99.33 {\scriptsize ± 0.54} & 66.67 {\scriptsize ± 27.22} & \gc \textbf{100.00} {\scriptsize ± 0.00} & 91.33 {\scriptsize ± 3.78} & 94.67 {\scriptsize ± 4.35} \\
synonyms & 10.00 {\scriptsize ± 4.50} & 22.67 {\scriptsize ± 5.62} & 25.00 {\scriptsize ± 8.83} & \gc \textbf{25.33} {\scriptsize ± 7.98} & 24.33 {\scriptsize ± 2.76} & 12.67 {\scriptsize ± 0.72} & 18.33 {\scriptsize ± 1.91} \\
taxonomy\_animal & 43.67 {\scriptsize ± 15.96} & 44.33 {\scriptsize ± 17.72} & 92.00 {\scriptsize ± 3.77} & 34.00 {\scriptsize ± 15.08} & 69.00 {\scriptsize ± 24.10} & 73.67 {\scriptsize ± 8.09} & \gc \textbf{99.67} {\scriptsize ± 0.27} \\
translation\_en-de & 84.67 {\scriptsize ± 1.19} & 74.00 {\scriptsize ± 3.30} & 85.33 {\scriptsize ± 0.72} & 77.33 {\scriptsize ± 2.60} & 83.67 {\scriptsize ± 1.19} & 57.00 {\scriptsize ± 20.82} & \gc \textbf{85.67} {\scriptsize ± 0.54} \\
translation\_en-es & \gc \textbf{90.67} {\scriptsize ± 0.98} & 83.33 {\scriptsize ± 3.07} & 88.33 {\scriptsize ± 1.78} & 83.67 {\scriptsize ± 2.76} & 89.00 {\scriptsize ± 0.47} & 85.33 {\scriptsize ± 0.27} & 86.00 {\scriptsize ± 2.05} \\
translation\_en-fr & 87.33 {\scriptsize ± 0.72} & 82.00 {\scriptsize ± 0.94} & \gc \textbf{88.00} {\scriptsize ± 1.63} & 84.00 {\scriptsize ± 2.05} & 87.67 {\scriptsize ± 1.91} & 84.67 {\scriptsize ± 3.14} & 83.00 {\scriptsize ± 2.36} \\
word\_sorting & 54.00 {\scriptsize ± 15.41} & 39.67 {\scriptsize ± 12.11} & 27.33 {\scriptsize ± 7.37} & \gc \textbf{71.00} {\scriptsize ± 4.50} & 54.00 {\scriptsize ± 15.06} & 36.33 {\scriptsize ± 11.49} & 53.33 {\scriptsize ± 8.38} \\
word\_unscrambling & 28.00 {\scriptsize ± 4.78} & 38.00 {\scriptsize ± 3.74} & 42.33 {\scriptsize ± 8.59} & 23.00 {\scriptsize ± 9.57} & \gc \textbf{52.00} {\scriptsize ± 7.79} & 43.00 {\scriptsize ± 1.25} & 48.00 {\scriptsize ± 7.59} \\
\midrule
\textbf{\# best-performing tasks} & 3 & 1 & 6 & 7 & 8 & 2 & \gc \textbf{18} \\
\textbf{Average Rank} & 4.10 & 4.97 & 3.60 & 4.50 & 2.90 & 4.77 & \gc \textbf{1.97} \\
\bottomrule
\end{tabular}
\end{adjustbox}
\label{tab:mainresults_full}
\vspace{-0.05em}
\end{table*}
We present experimental results on 30 instruction induction tasks in Table~\ref{tab:mainresults_full}. 
All methods are evaluated under the same settings using three different random seeds, and we report the average performance along with the standard error. 
Our proposed method, PRESTO, achieves the best performance on 18 out of 30 tasks, with an average rank of 1.97. 
This is more than twice the number of first-place finishes compared to the second-best method, ZOPO~\cite{hu2024localized}, which ranks first on 8 tasks and has an average rank of 2.90. 
These results indicate that PRESTO is not only effective on a few specific tasks but also demonstrates strong generalization across a wide range of tasks.
\section{NeuralUCB}
\label{supp_sec:neuralucb}
Here, we introduce the details about the NeuralUCB~\cite{zhou2020neural}.
We follow the overall architecture and hyperparameters used in~\cite{lin2024use}.
At each optimization step, the score predictor $m(g(z); \theta)$ is trained on previously evaluated soft prompts and their corresponding scores.
The model's predicted score $\mu(z)$ and its associated uncertainty $\sigma(z)$ are computed as:
\begin{align}
    \mu(z) &= m(g(z); \theta), \\
    \sigma(z) &= \sqrt{
        \nabla_\theta m(g(z); \theta)^\top V^{-1} \nabla_\theta m(g(z); \theta)
    }, \\
    \text{where} \quad V &= \sum_{\tau=1}^{t} \nabla_\theta m(g(z_\tau); \theta) \nabla_\theta m(g(z_\tau); \theta)^\top + \lambda I,
\end{align}
where $\lambda$ is a regularization coefficient and $t$ is the number of observed data.
The next prompt to evaluate is selected by maximizing an Upper Confidence Bound (UCB):
\begin{equation}
    z_{\text{next}} = \underset{z \in Z}{\arg\max}\  \mu(z) + \beta^{1/2} \sigma(z),
\end{equation}
where $\beta$ is a weighting parameter that balances exploration and exploitation.
Following~\cite{lin2024use}, $\lambda$ is set to 0.1 and $\beta$ is set to 1. 
The score predictor $m(\cdot; \theta)$ is a simple MLP with a hidden layer size is 100 and an output dimension is 1.
We use the Adam optimizer to train the MLP, and the learning rate is set to 0.001.
\section{Full Experimental Results of the Ablation Study}
\label{supp_sec:ablation}
In this section, we provide the full experimental results of the ablation study in Table~\ref{supp_table:ablation}.
The ablation study was performed over 20 instruction induction tasks, which are used in Table 1 of the main paper.
Starting from the vanilla method, we incrementally add the score sharing method, score consistency regularization method, and preimage-based initialization method.
Our proposed method, PRESTO, is the full model with all these components.
As shown in the Table, the performance consistently improves as each component is added sequentially. 
Notably, the model that incorporates all proposed modules achieves the best overall performance. 
These results demonstrate that each of the three modules we propose contributes meaningfully to the overall performance gain.
\begin{table}[htbp]
\centering
\caption{Ablation study of each component in 20 tasks. All experimental results are the mean and standard error of 3 different seeds.}
\label{supp_table:ablation}
\begin{adjustbox}{width=\textwidth}
\begin{tabular}{llllll}
\toprule
 \textbf{Tasks} &  \textbf{Vanilla} &  \textbf{+ SS} &  \textbf{+ SS, Reg} &  \textbf{+ SS, Init} &  \textbf{+ SS, Init, Reg} \\
\midrule 
antonyms                  & 80.00 {\scriptsize ± 0.82} & 81.33 {\scriptsize ± 2.37} & 82.67 {\scriptsize ± 0.54} & 79.67 {\scriptsize ± 2.88} & \gc \textbf{83.33} {\scriptsize ± 1.19} \\
auto\_categorization      & 17.00 {\scriptsize ± 1.63} & 24.67 {\scriptsize ± 4.72} & 28.67 {\scriptsize ± 3.03} & 31.00 {\scriptsize ± 0.82} & \gc \textbf{31.67} {\scriptsize ± 3.41} \\
auto\_debugging           & 10.00 {\scriptsize ± 4.71} & 12.50 {\scriptsize ± 5.89} & 8.33 {\scriptsize ± 6.80} & 20.00 {\scriptsize ± 0.00} & \gc \textbf{20.83} {\scriptsize ± 3.40} \\
cause\_and\_effect        & 74.67 {\scriptsize ± 14.28} & 87.00 {\scriptsize ± 0.47} & 93.33 {\scriptsize ± 3.93} & 93.33 {\scriptsize ± 2.88} & \gc \textbf{94.67} {\scriptsize ± 2.88} \\
common\_concept           & 19.44 {\scriptsize ± 2.21} & 18.24 {\scriptsize ± 1.79} & \gc \textbf{24.39} {\scriptsize ± 1.16} & 21.40 {\scriptsize ± 0.33} & 22.86 {\scriptsize ± 3.27} \\
diff                      & 81.00 {\scriptsize ± 2.16} & 87.00 {\scriptsize ± 4.24} & 97.00 {\scriptsize ± 0.82} & 93.67 {\scriptsize ± 4.01} & \gc \textbf{98.00} {\scriptsize ± 0.82} \\
informal\_to\_formal      & 50.67 {\scriptsize ± 3.11} & 51.96 {\scriptsize ± 5.16} & \gc \textbf{58.06} {\scriptsize ± 3.24} & 54.25 {\scriptsize ± 2.20} & 52.77 {\scriptsize ± 5.46} \\
letters\_list             & 98.33 {\scriptsize ± 1.36} & 99.67 {\scriptsize ± 0.27} & 99.67 {\scriptsize ± 0.27} & \gc \textbf{100.00} {\scriptsize ± 0.00} & 99.33 {\scriptsize ± 0.54} \\
negation                  & 76.33 {\scriptsize ± 1.91} & 85.33 {\scriptsize ± 1.66} & 84.33 {\scriptsize ± 3.03} & \gc \textbf{86.33} {\scriptsize ± 0.27} & 84.00 {\scriptsize ± 2.16} \\
object\_counting          & 40.67 {\scriptsize ± 10.14} & 39.67 {\scriptsize ± 3.47} & 44.33 {\scriptsize ± 3.78} & 43.67 {\scriptsize ± 0.98} & \gc \textbf{45.67} {\scriptsize ± 4.38} \\
odd\_one\_out             & 52.67 {\scriptsize ± 10.89} & 61.33 {\scriptsize ± 6.28} & 66.67 {\scriptsize ± 0.54} & 68.67 {\scriptsize ± 1.96} & \gc \textbf{70.00} {\scriptsize ± 0.94} \\
orthography\_starts\_with & 49.33 {\scriptsize ± 3.54} & 54.33 {\scriptsize ± 2.84} & 47.67 {\scriptsize ± 3.81} & 55.67 {\scriptsize ± 3.54} & \gc \textbf{57.33} {\scriptsize ± 6.08} \\
rhymes                    & 73.67 {\scriptsize ± 9.16} & 87.33 {\scriptsize ± 8.30} & \gc \textbf{96.00} {\scriptsize ± 1.70} & 86.00 {\scriptsize ± 5.91} & 85.00 {\scriptsize ± 7.41} \\
second\_word\_letter      & 49.67 {\scriptsize ± 18.16} & \gc \textbf{81.67} {\scriptsize ± 9.10} & 76.98 {\scriptsize ± 16.74} & 53.33 {\scriptsize ± 16.15} & 77.00 {\scriptsize ± 12.57} \\
sentence\_similarity      & 17.33 {\scriptsize ± 3.54} & \gc \textbf{22.67} {\scriptsize ± 3.57} & 17.33 {\scriptsize ± 4.46} & 21.33 {\scriptsize ± 5.30} & 21.67 {\scriptsize ± 8.49} \\
sum                       & 80.00 {\scriptsize ± 15.92} & 95.33 {\scriptsize ± 1.91} & \gc \textbf{96.67} {\scriptsize ± 2.72} & 95.67 {\scriptsize ± 3.54} & 94.67 {\scriptsize ± 4.35} \\
synonyms                  & 16.33 {\scriptsize ± 2.37} & \gc \textbf{19.00} {\scriptsize ± 0.47} & 15.67 {\scriptsize ± 3.57} & 18.33 {\scriptsize ± 2.13} & 18.33 {\scriptsize ± 1.91} \\
taxonomy\_animal          & 77.67 {\scriptsize ± 17.83} & 82.00 {\scriptsize ± 1.63} & 98.00 {\scriptsize ± 1.63} & 98.67 {\scriptsize ± 0.72} & \gc \textbf{99.67} {\scriptsize ± 0.27} \\
word\_sorting             & 24.00 {\scriptsize ± 0.47} & 53.67 {\scriptsize ± 15.78} & 46.00 {\scriptsize ± 11.09} & \gc \textbf{54.67} {\scriptsize ± 4.84} & 53.33 {\scriptsize ± 8.38} \\
word\_unscrambling        & 49.33 {\scriptsize ± 6.42} & 46.67 {\scriptsize ± 5.97} & 53.67 {\scriptsize ± 4.48} & \gc \textbf{60.67} {\scriptsize ± 0.72} & 48.00 {\scriptsize ± 7.59} \\
\midrule
\textbf{\# best-performing tasks} & 0 & 3 & 4 & 4 & \gc \textbf{9} \\
\textbf{Average Rank} & 4.55 & 3.10 & 2.65 & 2.30 & \gc \textbf{2.20} \\
\bottomrule
\end{tabular}
\end{adjustbox}
\end{table}

\section{Efficiency Analysis}
\label{supp_sec:efficiency}
Table~\ref{supp_tab:efficiency} summarizes the computation time required for each stage of our method.
We provide the means and standard errors over 30 tasks.
The preimage-based initialization step, computed only at the beginning of the optimization process, is notably efficient, taking only 27.67 ± 3.75 seconds on average.
Training the MLP model is also efficient, with the non-regularized version requiring just 1.52 ± 0.18 seconds to train the MLP at each iteration and the regularized variant taking 2.17 ± 0.20 seconds.
These results indicate that incorporating score consistency regularization introduces only a marginal overhead while potentially improving optimization performance, as shown in Table~\ref{supp_table:ablation}.
The overall total optimization process completes in 637.51 ± 81.66 seconds. 
Considering the complexity of the task, this runtime demonstrates that our method is computationally efficient and practical for real-world applications. 
\begin{table}[htbp]
\centering
\caption{Computation Time Summary}
\label{supp_tab:efficiency}
\begin{tabular}{ll}
\toprule
\textbf{Stage} & \textbf{Time (sec)} \\
\midrule
% Preimage generation & 8.06 $\pm$ 1.03 min \\
Preimage-based initialization & 27.67 $\pm$ 3.75 \\
\midrule
MLP train (w/o Regularization) & 1.52 $\pm$ 0.18 \\
MLP train (w/ Regularization) & 2.17 $\pm$ 0.20 \\
\midrule
Total optimization & 637.51 $\pm$ 81.66 \\
\bottomrule
\end{tabular}
\end{table}
% \begin{table}[htbp]
% \centering
% \caption{Computation Time Summary (in seconds)}
% \begin{tabular}{lS[table-format=3.2(2)]}
% \toprule
% \textbf{Stage} & \textbf{Time} \\
% \midrule
% \addlinespace[0.3em]
% \multicolumn{2}{l}{\textit{Initialization}} \\
% \hspace{1em}Preimage-based initialization & 27.67 \pm 20.56 \\
% \addlinespace[0.3em]
% \multicolumn{2}{l}{\textit{Training}} \\
% \hspace{1em}MLP (w/o Regularization) & 1.52 \pm 0.18 \\
% \hspace{1em}MLP (w/ Regularization) & 2.17 \pm 0.20 \\
% \addlinespace[0.3em]
% \multicolumn{2}{l}{\textit{Optimization}} \\
% \hspace{1em}Total optimization & 637.51 \pm 81.66 \\
% \bottomrule
% \end{tabular}
% \end{table}
\section{Computational Analysis of MMD}
\label{supp_sec:MMD_cost}

\begin{table}[htbp]
\centering
\caption{MMD Computation Time for Different Candidate Set Sizes}
\label{supp_tab:MMD_cost}
\begin{tabular}{ll}
\toprule
\textbf{Candidate Set Size} & \textbf{MMD Computation Time (sec)} \\
\midrule
1k & 7.18 $\pm$ 1.39 \\
5k & 11.13 $\pm$ 1.77 \\
10k (Current) & 27.67 $\pm$ 3.75 \\
20k & 79.92 $\pm$ 5.32 \\
30k & 94.31 $\pm$ 8.83 \\
\bottomrule
\end{tabular}
\end{table}
In table~\ref{supp_tab:MMD_cost}, we conducted a computational analysis of the MMD with respect to the size of the candidate set (1k, 5k, 10k, 20k, 30k).
As expected, the computation time increases with the size of the candidate set. 
Notably, even the largest setting (30k) remains computationally feasible, taking approximately 1 minute and 30 seconds.
In practice, we set the size of the candidate set as 10k accross all the tasks, which takes only 27.67 seconds.
\section{Effect of preimage size}
\label{supp_sec:preimage_size}
\begin{table}[htbp]
\centering
\caption{Average Accuracy for Different Preimage Sizes}
\label{supp_tab:preimage_accuracy}
\begin{tabular}{ll}
\toprule
\textbf{Preimage Size (\%)} & \textbf{Average Accuracy} \\
\midrule
0 (Vanilla) & 51.91 \\
1 & 59.95 \\
10 & 60.67 \\
50 & 61.89 \\
100 (Current) & 62.91 \\
\bottomrule
\end{tabular}
\end{table}
In table~\ref{supp_tab:preimage_accuracy}, we analyzed the effect of preimage size by varying the proportion of soft prompts included in each preimage (0\%, 1\%, 10\%, 50\%, and 100\%). 
The 0\% denotes the vanilla model, which does not leverage the preimage structure. 
The results show that larger preimage sizes lead to higher average accuracy, indicating that richer information in the preimage facilitates more successful optimization. 
This highlights the critical role of the preimage structure in instruction optimization.
\section{Computational Analysis of Preimage Construction}
\label{supp_sec:preimage_construction}

\begin{table}[htbp]
\centering
\caption{Preimage Construction Time and Memory Usage for Different Candidate Set Sizes}
\label{supp_tab:preimage_cost}
\begin{tabular}{lll}
\toprule
\textbf{Candidate Set Size} & \textbf{Preimage Construction Time (min.)} & \textbf{Memory (MB)} \\
\midrule
1k & 0.66 $\pm$ 0.04 & 47.47 \\
5k & 3.31 $\pm$ 0.20 & 237.46 \\
10k (Current) & 6.72 $\pm$ 0.39 & 474.96 \\
20k & 13.36 $\pm$ 0.80 & 950.10 \\
30k & 19.82 $\pm$ 1.02 & 1425.63 \\
\bottomrule
\end{tabular}
\end{table}
In table~\ref{supp_tab:preimage_cost}, we provide the computational cost analysis of preimage construction. 
The preimage construction time and memory usage for different candidate set sizes are as follows: for 1k candidates, 0.66 ± 0.04 minutes and 47.47 MB; for 5k candidates, 3.31 ± 0.20 minutes and 237.46 MB; for 10k candidates (current setting), 6.72 ± 0.39 minutes and 474.96 MB; for 20k candidates, 13.36 ± 0.80 minutes and 950.10 MB; and for 30k candidates, 19.82 ± 1.02 minutes and 1425.63 MB.
The results show that construction time and total memory usage increase approximately linearly with the size of the candidate set, while the overall cost remains modest. 

\begin{table}[htbp]
\centering
\caption{Preimage Construction Time for Different Numbers of Soft Prompt Tokens}
\label{supp_tab:softprompt_preimage}
\begin{tabular}{ll}
\toprule
\textbf{\# Soft Prompt Tokens} & \textbf{Preimage Construction Time (min.)} \\
\midrule
3 & 6.72 $\pm$ 0.39 \\
5 & 6.87 $\pm$ 0.44 \\
10 & 7.08 $\pm$ 0.45 \\
50 & 8.66 $\pm$ 0.62 \\
100 & 10.52 $\pm$ 0.70 \\
\bottomrule
\end{tabular}
\end{table}
We provide a scalability analysis of preimage construction with respect to the size of the soft prompt space, which is defined as (number of tokens $\times$ dimension) in table~\ref{supp_tab:softprompt_preimage}. 
Since the dimension is fixed (it depends on the white-box LLM), we focus on the number of soft prompt tokens: 3, 5, 10, 50, and 100. 
The table above shows that the proposed method has good scalability. 
With a large number of soft prompts (50 and 100), the preimage construction remains computationally feasible. 
In our experiments, we used 3 to 10 soft prompt tokens.
\section{Wall-clock time comparison}
\label{supp_sec:wall_clock_comp}

\begin{table}[htbp]
\centering
\caption{Comparison of Different Methods}
\label{supp_tab:method_comparison}
\begin{tabular}{lllll}
\toprule
 & \textbf{InstructZero~\cite{chen2024instructzero}} & \textbf{INSTINCT~\cite{lin2024use}} & \textbf{ZOPO~\cite{hu2024localized}} & \textbf{PRESTO (Ours)} \\
\midrule
Preprocess (min.) & - & 2.02 $\pm$ 0.38 & 5.07 $\pm$ 0.48 & 6.81 $\pm$ 0.51 \\
Optimization (min.) & 11.17 $\pm$ 1.04 & 13.21 $\pm$ 1.79 & 9.53 $\pm$ 1.19 & 10.63 $\pm$ 1.36 \\
Average Accuracy & 61.67 & 67.92 & 69.79 & \textbf{72.76} \\
\bottomrule
\end{tabular}
\end{table}
We conducted a wall-clock time comparison with baselines as provided in table~\ref{supp_tab:method_comparison}. 
During preprocessing, which is performed before the optimization process begins, PRESTO generates both LLM embeddings and instructions, whereas INSTINCT generates only the embeddings.
Despite involving more components, PRESTO achieves a lower overall optimization time than INSTINCT.
This is because PRESTO pre-generates instructions in batch during preprocessing, while INSTINCT queries the LLM at every optimization step.
As shown above, PRESTO incurs only marginal preprocessing overhead, yet achieves superior optimization performance.
\section{Impact of Hyperparameters on Preimage Structure}
\label{supp_sec:Preimage_hyperparameter}
\begin{figure}[t]
    \centering
    \includegraphics[width=\textwidth]{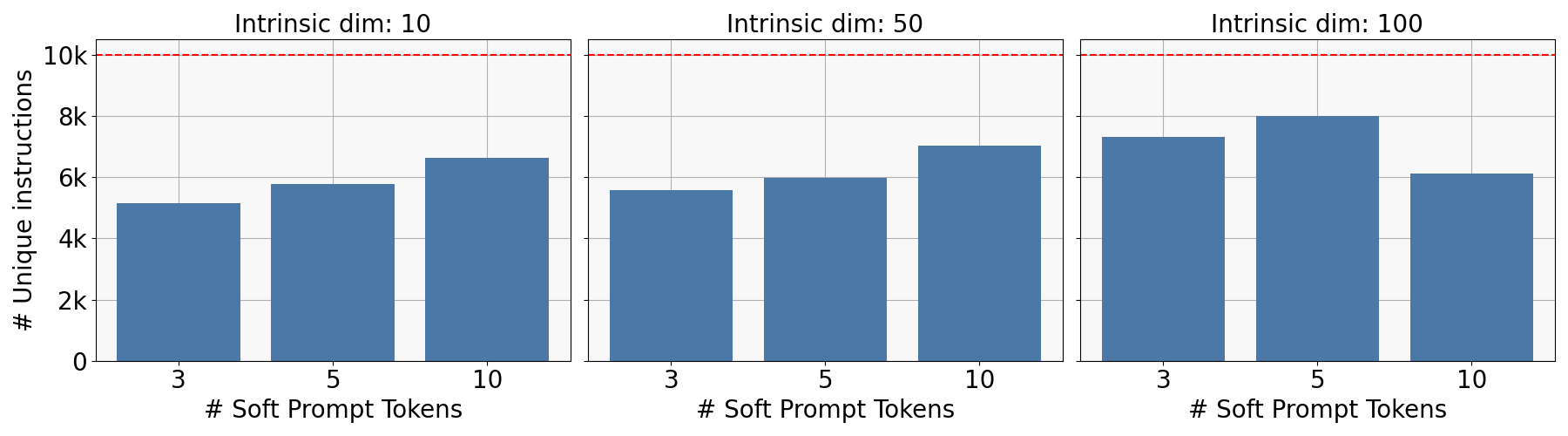}
    \caption{
    Impact of the intrinsic dimension and the number of soft prompt tokens on the preimage structure.
    } 
    \label{fig:hyperparameter_preimage}
    \vspace{-1.3em}
\end{figure}

Here, we present the impact of hyperparameters on preimage structure.
As demonstrated in prior work~\cite{lin2024use}, the intrinsic dimension has a direct impact on the distance between sampled soft prompts, significantly affecting the diversity of generated instructions.
In this study, we analyze the structure of the preimage with respect to the intrinsic dimension and the number of soft prompt tokens, both of which are key factors influencing performance.
As shown in Figure~\ref{fig:hyperparameter_preimage}, increasing the intrinsic dimension from 10 to 100 leads to a larger number of unique instructions. 
However, even at an intrinsic dimension of 100, a considerable number of duplicate instructions remain.

\section{Preimage Structures in Different White-box LLMs}
\label{supp_sec:preimage_whitebox}
\begin{figure}[t]
    \centering
    \begin{subfigure}[h]{0.48\linewidth}
        \hspace{1.8em}
        \includegraphics[width=0.7\textwidth]{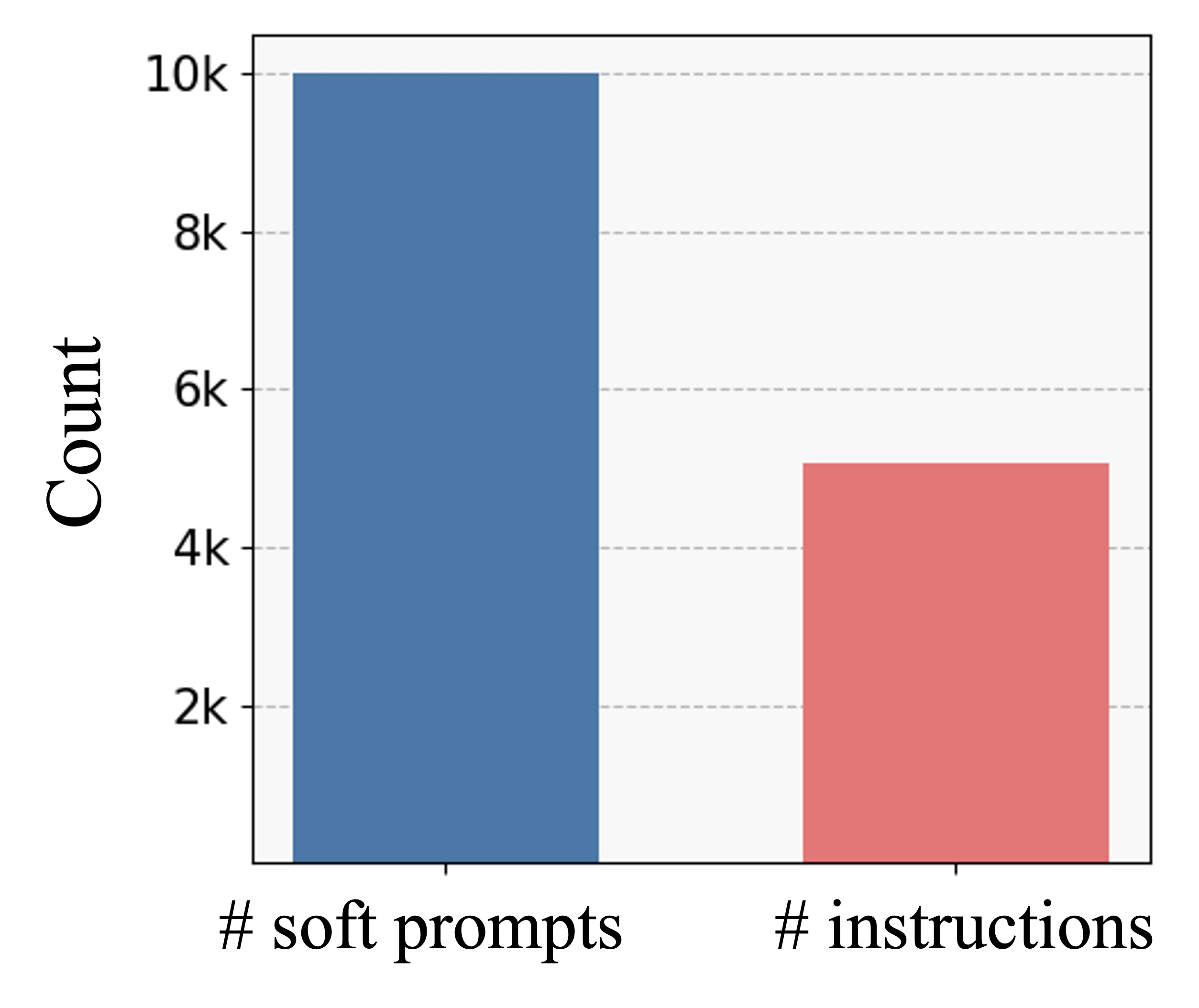}
        \caption{Results in Mistral-7B-Instruct-v0.3} 
    \end{subfigure}
    \begin{subfigure}[h]{0.48\linewidth}
        \hspace{2.0em}
        \includegraphics[width=0.7\textwidth]{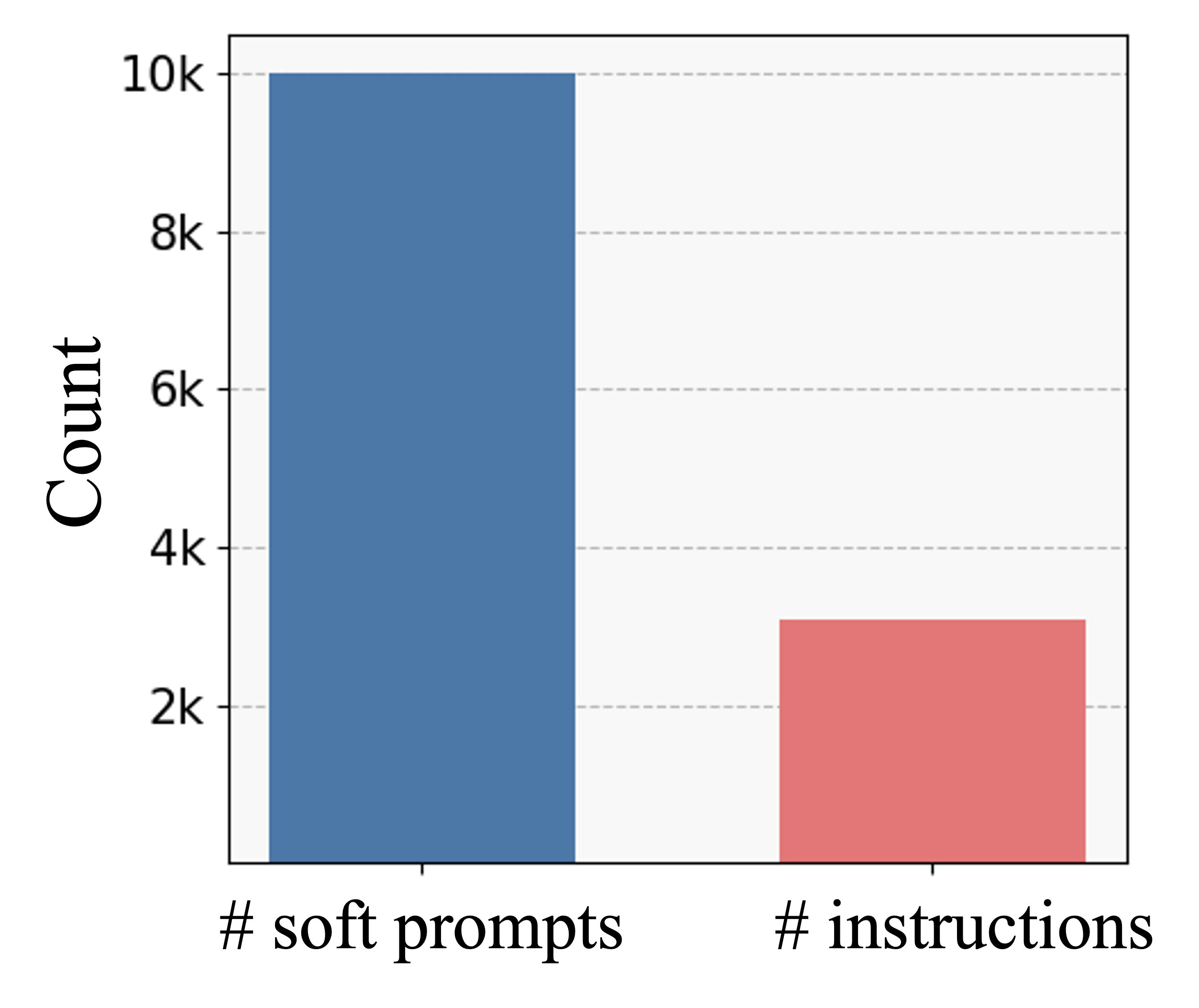}
        \caption{Results in Qwen2.5-7B-Instruct} 
    \end{subfigure}
    \caption{
    Number of unique instructions generated by 10,000 soft prompts in different white-box LLMs.
    }
    \label{fig:preimage_different_whitebox}
\end{figure}

In this section, we provide an additional analysis of the preimage structure under identical conditions using different white-box LLMs.
While the main experiments utilized LLaMA-3.1-8B-Instruct~\cite{grattafiori2024llama} and revealed a high degree of instruction duplication when sampling $N$ soft prompts at random, this section visualizes the preimage structures obtained from Mistral-7B-Instruct-v0.3~\cite{jiang2023mistral7b} and Qwen2.5-7B-Instruct\cite{yang2024qwen2} under the same sampling procedure.
The results represent averages across 30 instruction induction tasks, considering all combinations of three intrinsic dimensions, $[10, 50, 100]$, and three soft prompt token numbers, $[3,5,10]$.
As shown in Figure~\ref{fig:preimage_different_whitebox}, Mistral generated approximately 50$\%$ duplicate instructions when sampling 10,000 soft prompts, while Qwen produced fewer than 3,000 unique instructions under the same conditions.
\section{Different Combinations of White-box LLMs and Black-box LLMs}
\label{supp_sec:LLMComb}
\begin{table}[t]
\centering
\caption{Instruction optimization results for GPT-4.1 with various white-box LLMs. We omit LLaMA to avoid redundancy, as it is already reported in Table~\ref{supp_table:ablation}.}
\begin{adjustbox}{width=0.7\textwidth}
\begin{tabular}{lll||ll}
\toprule
\textbf{Black-box LLM} & \textbf{GPT4.1} & \textbf{GPT4.1} & \textbf{GPT4.1} & \textbf{GPT4.1} \\
\textbf{White-box LLM} & \textbf{Qwen} & \textbf{Qwen} & \textbf{Mistral} & \textbf{Mistral} \\
\hline
\textbf{Tasks} & \textbf{Vanilla} & \textbf{PRESTO} & \textbf{Vanilla} & \textbf{PRESTO} \\
\hline
antonyms                  & 78.00  & 85.00  {\scriptsize \textcolor{blue}{(+7.00)}} & 83.00  & 87.00  {\scriptsize \textcolor{blue}{(+4.00)}} \\
auto\_categorization       & 24.00  & 33.00  {\scriptsize \textcolor{blue}{(+9.00)}} & 29.00  & 36.00  {\scriptsize \textcolor{blue}{(+7.00)}} \\
auto\_debugging           & 0.00   & 0.00   {\scriptsize \textcolor{gray}{(+0.00)}} & 0.00   & 25.00  {\scriptsize \textcolor{blue}{(+25.00)}} \\
cause\_and\_effect        & 92.00  & 92.00  {\scriptsize \textcolor{gray}{(+0.00)}} & 92.00  & 92.00  {\scriptsize \textcolor{gray}{(+0.00)}} \\
common\_concept           & 28.86  & 32.85  {\scriptsize \textcolor{blue}{(+3.99)}} & 25.51  & 30.14  {\scriptsize \textcolor{blue}{(+4.63)}} \\
diff                      & 96.00  & 100.00 {\scriptsize \textcolor{blue}{(+4.00)}} & 85.00  & 93.00  {\scriptsize \textcolor{blue}{(+8.00)}} \\
informal\_to\_formal      & 59.86  & 63.85  {\scriptsize \textcolor{blue}{(+3.99)}} & 61.45  & 53.70  {\scriptsize \textcolor{red}{(-7.75)}} \\
letters\_list             & 100.00 & 98.00  {\scriptsize \textcolor{red}{(-2.00)}}  & 100.00 & 100.00 {\scriptsize \textcolor{gray}{(+0.00)}} \\
negation                  & 82.00  & 82.00  {\scriptsize \textcolor{gray}{(+0.00)}} & 81.00  & 81.00  {\scriptsize \textcolor{gray}{(+0.00)}} \\
object\_counting          & 31.00  & 34.00  {\scriptsize \textcolor{blue}{(+3.00)}} & 26.00  & 54.00  {\scriptsize \textcolor{blue}{(+28.00)}} \\
odd\_one\_out             & 68.00  & 74.00  {\scriptsize \textcolor{blue}{(+6.00)}} & 66.00  & 72.00  {\scriptsize \textcolor{blue}{(+6.00)}} \\
orthography\_starts\_with & 69.00  & 70.00  {\scriptsize \textcolor{blue}{(+1.00)}} & 34.00  & 34.00  {\scriptsize \textcolor{gray}{(+0.00)}} \\
rhymes                    & 4.00   & 59.00  {\scriptsize \textcolor{blue}{(+55.00)}}& 72.00  & 72.00  {\scriptsize \textcolor{gray}{(+0.00)}} \\
second\_word\_letter      & 82.00  & 100.00 {\scriptsize \textcolor{blue}{(+18.00)}}& 10.00  & 98.00  {\scriptsize \textcolor{blue}{(+88.00)}} \\
sentence\_similarity      & 30.00  & 26.00  {\scriptsize \textcolor{red}{(-4.00)}}  & 15.00  & 17.00  {\scriptsize \textcolor{blue}{(+2.00)}} \\
sum                       & 97.00  & 97.00  {\scriptsize \textcolor{gray}{(+0.00)}} & 100.00 & 100.00 {\scriptsize \textcolor{gray}{(+0.00)}} \\
synonyms                  & 25.00  & 39.00  {\scriptsize \textcolor{blue}{(+14.00)}}& 38.00  & 47.00  {\scriptsize \textcolor{blue}{(+9.00)}} \\
taxonomy\_animal          & 69.00  & 81.00  {\scriptsize \textcolor{blue}{(+12.00)}}& 81.00  & 100.00 {\scriptsize \textcolor{blue}{(+19.00)}} \\
word\_sorting             & 70.00  & 80.00  {\scriptsize \textcolor{blue}{(+10.00)}}& 78.00  & 77.00  {\scriptsize \textcolor{red}{(-1.00)}} \\
word\_unscrambling        & 45.00  & 47.00  {\scriptsize \textcolor{blue}{(+2.00)}} & 48.00  & 48.00  {\scriptsize \textcolor{gray}{(+0.00)}} \\
\bottomrule
\end{tabular}
\label{tab:llm_comparison_gpt}
\end{adjustbox}
\end{table}
\begin{table}[ht]
\caption{Performance of Gemini-2.0-flash with various white-box LLMs. For brevity, we write Gemini-2.0-flash as Gemini-2.0-f.}
\centering
\begin{adjustbox}{width=\textwidth}
\begin{tabular}{lll||ll||ll}
\toprule
\textbf{Black-box LLM} & \textbf{Gemini-2.0-f.} & \textbf{Gemini-2.0-f.} & \textbf{Gemini-2.0-f.} & \textbf{Gemini-2.0-f.} & \textbf{Gemini-2.0-f.} & \textbf{Gemini-2.0-f.} \\
\textbf{White-box LLM} & \textbf{LLaMA} & \textbf{LLaMA} & \textbf{Qwen} & \textbf{Qwen} & \textbf{Mistral} & \textbf{Mistral} \\
\hline
\textbf{Tasks} & \textbf{Vanilla} & \textbf{PRESTO} & \textbf{Vanilla} & \textbf{PRESTO} & \textbf{Vanilla} & \textbf{PRESTO} \\
\hline
antonyms                  & 72.00  & 84.00 {\scriptsize \textcolor{blue}{(+12.00)}} & 70.00  & 85.00 {\scriptsize \textcolor{blue}{(+15.00)}} & 85.00  & 88.00 {\scriptsize \textcolor{blue}{(+3.00)}} \\
auto\_categorization      & 31.00  & 29.00 {\scriptsize \textcolor{red}{(-2.00)}}   & 20.00  & 34.00 {\scriptsize \textcolor{blue}{(+14.00)}} & 33.00  & 24.00 {\scriptsize \textcolor{red}{(-9.00)}} \\
auto\_debugging           & 0.00   & 12.50 {\scriptsize \textcolor{blue}{(+12.50)}} & 12.50  & 12.50 {\scriptsize \textcolor{gray}{(+0.00)}}  & 12.50  & 0.00  {\scriptsize \textcolor{red}{(-12.50)}} \\
cause\_and\_effect        & 88.00  & 88.00 {\scriptsize \textcolor{gray}{(+0.00)}}  & 88.00  & 88.00 {\scriptsize \textcolor{gray}{(+0.00)}}  & 92.00  & 100.00{\scriptsize \textcolor{blue}{(+8.00)}} \\
common\_concept           & 12.19  & 29.57 {\scriptsize \textcolor{blue}{(+17.38)}} & 26.47  & 30.31 {\scriptsize \textcolor{blue}{(+3.84)}}  & 25.31  & 20.00 {\scriptsize \textcolor{red}{(-5.31)}} \\
diff                      & 99.00  & 100.00{\scriptsize \textcolor{blue}{(+1.00)}}  & 100.00 & 100.00{\scriptsize \textcolor{gray}{(+0.00)}}  & 98.00  & 98.00 {\scriptsize \textcolor{gray}{(+0.00)}} \\
informal\_to\_formal      & 56.87  & 46.45 {\scriptsize \textcolor{red}{(-10.42)}}  & 58.18  & 57.22 {\scriptsize \textcolor{red}{(-0.96)}}   & 60.46  & 61.28 {\scriptsize \textcolor{blue}{(+0.82)}} \\
letters\_list             & 100.00 & 100.00{\scriptsize \textcolor{gray}{(+0.00)}}  & 100.00 & 100.00{\scriptsize \textcolor{gray}{(+0.00)}}  & 100.00 & 100.00{\scriptsize \textcolor{gray}{(+0.00)}} \\
negation                  & 80.00  & 80.00 {\scriptsize \textcolor{gray}{(+0.00)}}  & 81.00  & 85.00 {\scriptsize \textcolor{blue}{(+4.00)}}  & 82.00  & 82.00 {\scriptsize \textcolor{gray}{(+0.00)}} \\
object\_counting          & 56.00  & 59.00 {\scriptsize \textcolor{blue}{(+3.00)}}  & 39.00  & 56.00 {\scriptsize \textcolor{blue}{(+17.00)}} & 41.00  & 52.00 {\scriptsize \textcolor{blue}{(+11.00)}} \\
odd\_one\_out             & 76.00  & 76.00 {\scriptsize \textcolor{gray}{(+0.00)}}  & 76.00  & 76.00 {\scriptsize \textcolor{gray}{(+0.00)}}  & 70.00  & 70.00 {\scriptsize \textcolor{gray}{(+0.00)}} \\
orthography\_starts\_with & 40.00  & 67.00 {\scriptsize \textcolor{blue}{(+27.00)}} & 64.00  & 67.00 {\scriptsize \textcolor{blue}{(+3.00)}}  & 53.00  & 51.00 {\scriptsize \textcolor{red}{(-2.00)}} \\
rhymes                    & 96.00  & 96.00 {\scriptsize \textcolor{gray}{(+0.00)}}  & 19.00  & 79.00 {\scriptsize \textcolor{blue}{(+60.00)}} & 92.00  & 98.00 {\scriptsize \textcolor{blue}{(+6.00)}} \\
second\_word\_letter      & 36.00  & 84.00 {\scriptsize \textcolor{blue}{(+48.00)}} & 99.00  & 99.00 {\scriptsize \textcolor{gray}{(+0.00)}}  & 56.00  & 59.00 {\scriptsize \textcolor{blue}{(+3.00)}} \\
sentence\_similarity      & 0.00   & 12.00 {\scriptsize \textcolor{blue}{(+12.00)}} & 19.00  & 27.00 {\scriptsize \textcolor{blue}{(+8.00)}}  & 9.00   & 10.00 {\scriptsize \textcolor{blue}{(+1.00)}} \\
sum                       & 89.00  & 100.00{\scriptsize \textcolor{blue}{(+11.00)}} & 100.00 & 100.00{\scriptsize \textcolor{gray}{(+0.00)}}  & 97.00  & 99.00 {\scriptsize \textcolor{blue}{(+2.00)}} \\
synonyms                  & 18.00  & 38.00 {\scriptsize \textcolor{blue}{(+20.00)}} & 37.00  & 41.00 {\scriptsize \textcolor{blue}{(+4.00)}}  & 33.00  & 41.00 {\scriptsize \textcolor{blue}{(+8.00)}} \\
taxonomy\_animal          & 94.00  & 98.00 {\scriptsize \textcolor{blue}{(+4.00)}}  & 76.00  & 76.00 {\scriptsize \textcolor{gray}{(+0.00)}}  & 97.00  & 100.00{\scriptsize \textcolor{blue}{(+3.00)}} \\
word\_sorting             & 44.00  & 55.00 {\scriptsize \textcolor{blue}{(+11.00)}} & 75.00  & 78.00 {\scriptsize \textcolor{blue}{(+3.00)}}  & 50.00  & 73.00 {\scriptsize \textcolor{blue}{(+23.00)}} \\
word\_unscrambling        & 45.00  & 52.00 {\scriptsize \textcolor{blue}{(+7.00)}}  & 25.00  & 25.00 {\scriptsize \textcolor{gray}{(+0.00)}}  & 25.00  & 54.00 {\scriptsize \textcolor{blue}{(+29.00)}} \\
\bottomrule
\end{tabular}
\label{tab:gemini_comparison}
\end{adjustbox}
\end{table}

We evaluate the performance of our proposed method, PRESTO, as well as a vanilla variant that excludes its three core components: score sharing, preimage-based initialization, and score consistency regularization, across various combinations of white-box and black-box LLMs.
For white-box LLMs, we use LLaMa-3.1-8B-Instruct~\cite{grattafiori2024llama}, Qwen2.5-7B-Instruct~\cite{yang2024qwen2}, and Mistral-7B-Instruct-v0.3~\cite{jiang2023mistral7b}. 
As black-box LLMs, we use GPT-4.1 and Gemini-2.0-Flash.

Table~\ref{tab:llm_comparison_gpt} shows the performance using GPT-4.1 as the black-box LLM. 
We omit the results for the LLaMA here to avoid redundancy, as they are already reported in Table~\ref{supp_table:ablation}.
Both Qwen and Mistral show substantial performance improvements when PRESTO is applied. 
Notably, Qwen achieves a +55 gain on the rhymes task, while Mistral sees a +88 improvement on the second\_word\_letter task.
Table~\ref{tab:gemini_comparison} presents results for optimizing instructions for Gemini-2.0-Flash using all three white-box LLMs.
Again, we observe consistent improvements: LLaMA achieves a +48 gain on the second\_word\_letter task, Qwen improves by +60 on rhymes, and Mistral sees a +29 increase on word\_unscrambling.
\section{Impact of Hyperparameters in Score Consistency Regularization}
\label{supp_sec:reg_hyperparameter}
We provide an analysis of the hyperparameter sensitivity of the score consistency regularization.
To prevent the score predictor from converging to incorrect estimates too early in training, we employ a linear scheduling strategy defined as $\gamma(t) = \gamma_{\text{max}}\cdot \min (1, t/T)$.
We fix $\gamma_{\text{max}}$ as 0.1 and $T$ as half of the full training epoch, 500.
In Table~\ref{supp_table: Reg comparision}, we report performance under different values of $\gamma$ and with or without scheduling.
The results show that $\gamma=0.1$ with scheduling yields the best performance, while $\gamma=1.0$ also achieves competitive results.
This indicates that our score consistency regularization is relatively insensitive to the choice of $\gamma$.
However, $\gamma = 0.1$ without scheduling leads to the worst performance, suggesting that the score predictor can converge to incorrect predictions if scheduling is not applied.
\begin{table}[ht]
\centering
\caption{Performance Comparison: Effect of $\gamma$ and Scheduling}
\label{supp_table: Reg comparision}
\begin{adjustbox}{width=0.8\textwidth}
\begin{tabular}{llll}
\toprule
\textbf{Tasks} & $\gamma = 0.1$ & $\gamma = 1.0$ & $\gamma = 0.1$, No schedule \\
\midrule
antonyms                  & \gc \textbf{83.33} {\scriptsize ± 1.19} & 82.00 {\scriptsize ± 3.12} & 78.00 {\scriptsize ± 4.32} \\
auto\_categorization      & 31.67 {\scriptsize ± 3.41} & \gc \textbf{32.33} {\scriptsize ± 1.32} & 29.13 {\scriptsize ± 2.22} \\
auto\_debugging           & \gc \textbf{20.83} {\scriptsize ± 3.40} & 18.74 {\scriptsize ± 2.89} & 19.52 {\scriptsize ± 1.26} \\
cause\_and\_effect        & \gc \textbf{94.67} {\scriptsize ± 2.88} & 93.43 {\scriptsize ± 3.21} & 93.33 {\scriptsize ± 2.31} \\
common\_concept           & \gc \textbf{22.86} {\scriptsize ± 3.27} & 19.44 {\scriptsize ± 0.98} & 12.43 {\scriptsize ± 6.43} \\
diff                      & 98.00 {\scriptsize ± 0.82} & \gc \textbf{98.32} {\scriptsize ± 0.12} & 95.67 {\scriptsize ± 1.34} \\
informal\_to\_formal      & 52.77 {\scriptsize ± 5.46} & 54.74 {\scriptsize ± 0.53} & \gc \textbf{57.50} {\scriptsize ± 4.76} \\
letters\_list             & 99.33 {\scriptsize ± 0.54} & \gc \textbf{100.00} {\scriptsize ± 0.00} & 99.33 {\scriptsize ± 0.54} \\
negation                  & \gc \textbf{84.00} {\scriptsize ± 2.16} & 81.00 {\scriptsize ± 2.11} & 82.00 {\scriptsize ± 1.98} \\
object\_counting          & \gc \textbf{45.67} {\scriptsize ± 4.38} & 44.33 {\scriptsize ± 3.21} & 43.89 {\scriptsize ± 2.19} \\
odd\_one\_out             & \gc \textbf{70.00} {\scriptsize ± 0.94} & 67.67 {\scriptsize ± 1.53} & 69.00 {\scriptsize ± 1.22} \\
orthography\_starts\_with & 57.33 {\scriptsize ± 6.08} & \gc \textbf{65.00} {\scriptsize ± 5.82} & 64.00 {\scriptsize ± 4.50} \\
rhymes                    & 85.00 {\scriptsize ± 7.41} & \gc \textbf{89.00} {\scriptsize ± 8.12} & 87.00 {\scriptsize ± 3.42} \\
second\_word\_letter      & \gc \textbf{77.00} {\scriptsize ± 12.57} & 73.33 {\scriptsize ± 15.32} & 64.83 {\scriptsize ± 10.22} \\
sentence\_similarity      & 21.67 {\scriptsize ± 8.49} & \gc \textbf{22.33} {\scriptsize ± 9.34} & 19.67 {\scriptsize ± 7.89} \\
sum                       & 94.67 {\scriptsize ± 4.35} & \gc \textbf{95.00} {\scriptsize ± 3.90} & 94.32 {\scriptsize ± 1.90} \\
synonyms                  & \gc \textbf{18.33} {\scriptsize ± 1.91} & 17.67 {\scriptsize ± 0.91} & 16.33 {\scriptsize ± 3.01} \\
taxonomy\_animal          & \gc \textbf{99.67} {\scriptsize ± 0.27} & 97.33 {\scriptsize ± 0.87} & 96.67 {\scriptsize ± 0.12} \\
word\_sorting             & \gc \textbf{53.33} {\scriptsize ± 8.38} & 48.00 {\scriptsize ± 7.76} & 42.00 {\scriptsize ± 6.76} \\
word\_unscrambling        & 48.00 {\scriptsize ± 7.59} & 38.00 {\scriptsize ± 9.82} & \gc \textbf{52.00} {\scriptsize ± 7.69} \\
\bottomrule
\end{tabular}
\end{adjustbox}
\end{table}

\section{Best Instructions Discovered by PRESTO}
\label{supp_sec:instruction_example}
In Table~\ref{tab:best_instructions} and Table~\ref{tab:best_instructions2}, we provide the best instructions for each task found by our PRESTO.
For tasks like active\_to\_passive, cause\_and\_effect, and first\_word\_letter, PRESTO found instructions that directly command the black-box LLM to solve the task.
In contrast, for tasks like antonyms, auto\_categorization, and common\_concept, PRESTO found instructions by combining a command with in-context examples.
\begin{table}[htbp]
\caption{Best instructions for each task.}
\centering
\small
\renewcommand{\arraystretch}{1.2}
\begin{tabular}{>{\raggedright\arraybackslash}p{3cm}|>{\raggedright\arraybackslash}p{9cm}}
\hline
\textbf{Tasks} & \textbf{Best Instructions} \\
\hline
active\_to\_passive & to change the sentence structure to make the subject of the sentence the one receiving the action, rather than the one performing the action. \\
\hline
antonyms & It looks like you're trying to test my ability to find the opposite or complementary term for a given word. I'm happy to play along! Here are the answers: 1. humorless -> humorous 2. depressing -> cheerful 3. unwrap -> wrap 4. consumptive -> generative \\
\hline
auto\_categorization & It seems like you're providing a list of words and asking me to identify a common theme or category among them. For the first set of words: Input: Nature Nanotechnology, Annual Review of Biochemistry, and The Lancet Neurology Output: top journals This is correct! \\
\hline
auto\_debugging & I can help you with the following: Input: sum = 0for i in range(100, 0, -2): sum += i\ Output: 2550 Input: ``` class MyClass(): def init(self, param): self.param = param \\
\hline
cause\_and\_effect & The instruction was to determine the most likely cause-and-effect relationship between two sentences, and to output the sentence that is most likely to be the cause of the effect described in the other sentence. \\
\hline
common\_concept & Input: guitars, pendulums Output: involve oscillations. Input: snowman, marzipan fruit Output: inanimate, but made to imitate something animate. Input: ballet, snow Output: use specialized shoes.\\
\hline
diff & Based on the input-output pairs, it appears that the instruction is to subtract the second number from the first number. So, the instruction is: "Subtract the second number from the first number.” \\
\hline
first\_word\_letter & truncate the input to the first 1 character. \\
\hline
informal\_to\_formal & Input: The instructions were given. Output: The instructions were provided. Input: Do you want to go with me? Output: Shall I go with you? Input: I want to go with you. Output: I shall go with you. Input: I want to go with you. Output: I shall go with you. \\
\hline
larger\_animal & The instruction is to return the animal that is the largest in size among the two input animals. \\
\hline
letters\_list & split the input string into individual letters. \\
\hline
negation & Input: Cany Ash and Robert Sakula are both Architects. Output: Cany Ash and Robert Sakula are not both Architects. Input: The arket price of skill is reflected by wages. Output: The arket price of skill is not reflected by wages. \\
\hline
num\_to\_verbal & to convert numbers into words. The input is a number, and the output is the written form of that number. \\
\hline
object\_counting & Input: I have a watermelon, a cantaloupe, a honeydew, a watermelon, and a fruit salad. Output: 6 \\
\hline
odd\_one\_out & select the item that is not like the others. \\
\hline
orthography\_starts\_with & It seems like the instruction is to extract a word from a sentence based on a specific letter. The letter is indicated in square brackets at the end of the sentence. In that case, the instruction was to extract a word that starts with the letter indicated in the brackets. \\
\hline
periodic\_elements & return the name of the element with the given atomic number. \\
\hline
rhymes & reverse the order of the words in the input. \\
\hline
second\_word\_letter & to return the second letter of the input string. \\
\hline
sentence\_similarity &  I can help you with the task. It seems like you want me to determine the similarity between two sentences based on the context. If the sentences are about the same topic, you want to return 3 - probably. If they are about different topics, you want to return 0 - definitely not. \\
\hline
sentiment & The instruction was to identify the sentiment of each input as either positive or negative. \\
\hline
% singular\_to\_plural & pluralize the input noun. \\
% \hline
% sum & The instruction was to add the two numbers together and output the result. \\
% \hline
% synonyms & It seems like the instruction was to provide a list of word pairs with their corresponding synonyms. Here is the list: 1. propose - offer 2. probe - investigation 3. healthy - sound 4. spy - sight \\
% \hline
% taxonomy\_animal & The instruction was to remove the items that are not animals from the input lists. \\
% \hline
% translation\_en-de & The instruction was to translate the input into the corresponding output in the target language, which appears to be German. Here are the translations: 1. Input: label Output: etikettieren (or etikettieren, both are correct) 2. Input: emergency Output: Notstand \\
% \hline
% translation\_en-es & translate the input to Spanish. \\
% \hline
% translation\_en-fr & The instruction was to transform words into their French translations. \\
% \hline
% word\_sorting & I can solve this problem. The problem is to reorder the words in the list to be in alphabetical order. Input: List: discordant kilohm lulu Output: discordant kilohm lulu The list is already in alphabetical order. \\
% \hline
% word\_unscrambling & It appears that the input is a scrambled version of a word or phrase, and the output is the unscrambled version. \\
% \hline
\end{tabular}
\label{tab:best_instructions}
\end{table}
\begin{table}[htbp]
\caption{Best instructions for each task (continue).}
\centering
\small
\renewcommand{\arraystretch}{1.2}
\begin{tabular}{>{\raggedright\arraybackslash}p{3cm}|>{\raggedright\arraybackslash}p{9cm}}
\hline
\textbf{Tasks} & \textbf{Best Instructions} \\
\hline
singular\_to\_plural & pluralize the input noun. \\
\hline
sum & The instruction was to add the two numbers together and output the result. \\
\hline
synonyms & It seems like the instruction was to provide a list of word pairs with their corresponding synonyms. Here is the list: 1. propose - offer 2. probe - investigation 3. healthy - sound 4. spy - sight \\
\hline
taxonomy\_animal & The instruction was to remove the items that are not animals from the input lists. \\
\hline
translation\_en-de & The instruction was to translate the input into the corresponding output in the target language, which appears to be German. Here are the translations: 1. Input: label Output: etikettieren (or etikettieren, both are correct) 2. Input: emergency Output: Notstand \\
\hline
translation\_en-es & translate the input to Spanish. \\
\hline
translation\_en-fr & The instruction was to transform words into their French translations. \\
\hline
word\_sorting & I can solve this problem. The problem is to reorder the words in the list to be in alphabetical order. Input: List: discordant kilohm lulu Output: discordant kilohm lulu The list is already in alphabetical order. \\
\hline
word\_unscrambling & It appears that the input is a scrambled version of a word or phrase, and the output is the unscrambled version. \\
\hline
\end{tabular}
\label{tab:best_instructions2}
\end{table}
\section{Details of Preimage-based Initialization}
\label{supp_sec:preimage_details}
Here, we provide the details of the preimage-based initialization method.
To compute the representativeness score $S_{\text{rep}}$, we use the squared Maximum Mean Discrepancy (MMD$^2$), a widely used metric for measuring the similarity between two sets $X$ and $Y$~\cite{gretton2012kernel, arbel2019maximum, dziugaite2015training}.
For the kernel function in MMD, we adopt the Gaussian Radial Basis Function (RBF) kernel, $k(x, y) = \exp(-|| x-y|| / 2\sigma^2)$, where the bandwidth $\sigma$ determined using the commonly employed median heuristic: $\sigma = \text{median}\{\|u-v\| \ | \ u,v\in X\cup Y, u\ne v\}$~\cite{garreau2017large}.
We observed that preimages with a size less than 5 are rarely selected due to the influence of our size score $S_{\text{size}}$.
To reduce computational cost, we therefore consider only preimages with size greater than 4 during the preimage-based initialization.
\section{Instruction Generation Format}
\label{supp_sec:Instruction_format}
\begin{figure}[t]
    \centering
    \includegraphics[width=0.8\textwidth]{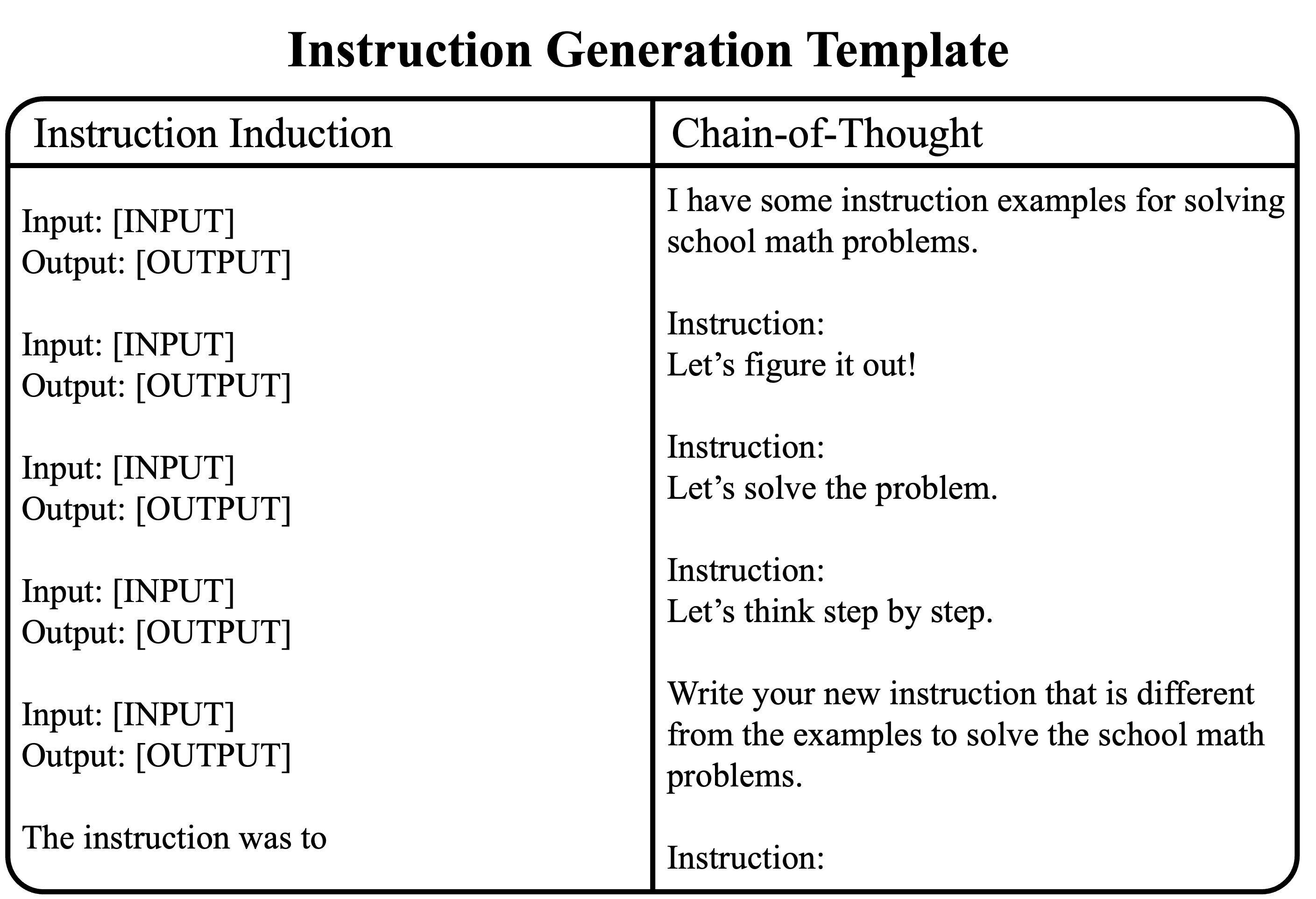}
    \caption{
    Instruction generation template for instruction induction task and chain-of-thought.
    } 
    \label{fig:instruction_generation_template}
\end{figure}

\begin{figure}[t]
    \centering
    \includegraphics[width=0.55\textwidth]{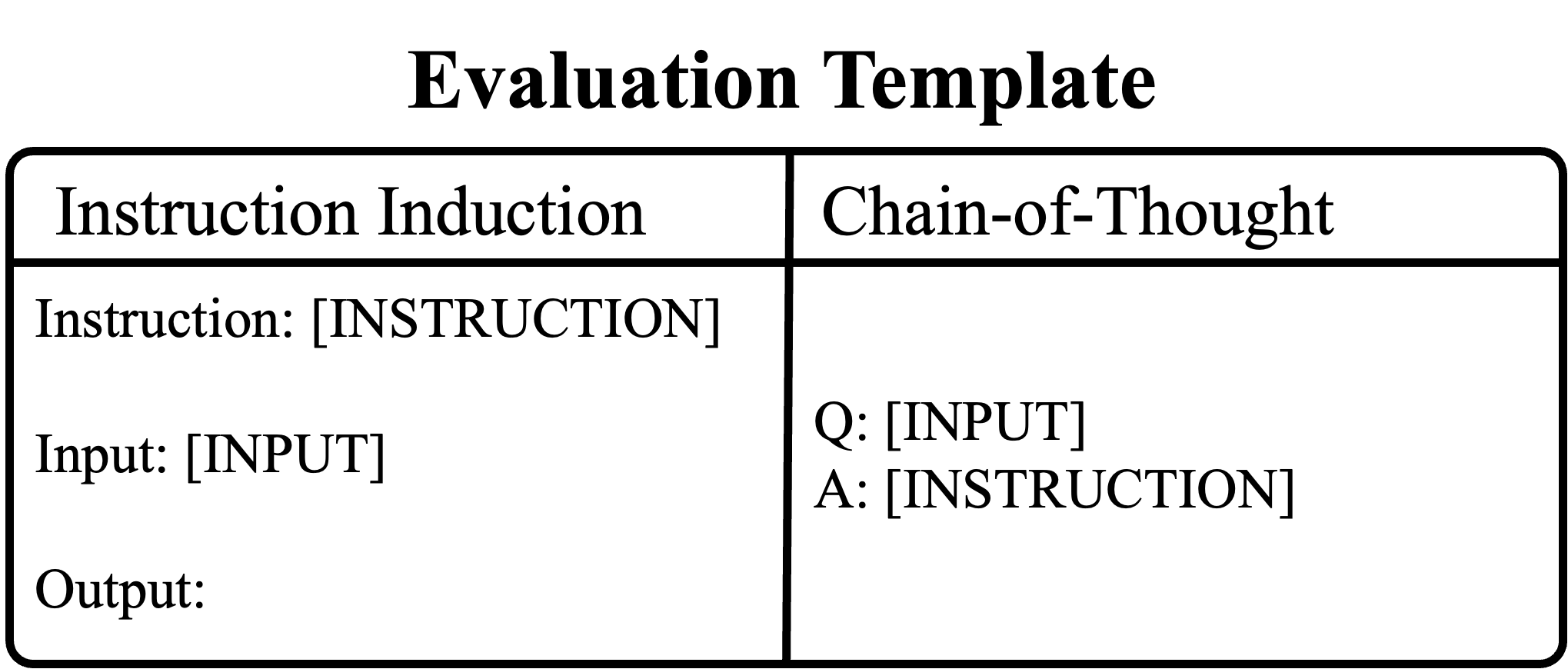}
    \caption{
    Evaluation template for instruction induction task and chain-of-thought.
    } 
    \label{fig:evaluation_tempalte}
\end{figure}

We present input templates for LLM-based instruction generation and evaluation in Figure~\ref{fig:instruction_generation_template} and Figure~\ref{fig:evaluation_tempalte}, respectively, covering both instruction induction and chain-of-thought tasks. 
For instruction induction tasks, we adopt the templates proposed in~\cite{chen2024instructzero}, and for chain-of-thought tasks, we utilize the templates introduced in~\cite{pryzant2023automatic}. 
In the instruction generation template (Figure~\ref{fig:instruction_generation_template}), each instance of [INPUT] and [OUTPUT] is replaced with a corresponding exemplar from a predefined exemplar set $E$. 
These exemplars remain fixed throughout the optimization process for a given task. 
During optimization, soft prompts are concatenated with the token embeddings of the instruction generation template (Figure~\ref{fig:instruction_generation_template}). 
The instruction produced from this template is then inserted into the [INSTRUCTION] slot of the evaluation template shown in Figure~\ref{fig:evaluation_tempalte}.

\end{document}